\documentclass{article}
\usepackage{etoolbox}
\usepackage{listings}

\usepackage{hyperref}
\usepackage[accepted]{icml2024}

\usepackage{amsmath,amssymb,amsfonts}
\usepackage{algorithmic}
\usepackage{graphicx}
\usepackage{subcaption}
\expandafter\def\csname ver@subfig.sty\endcsname{}

\usepackage{textcomp}
\usepackage{xcolor}
\usepackage{listings}
\definecolor{codegreen}{rgb}{0,0.6,0}
\definecolor{codegray}{rgb}{0.5,0.5,0.5}
\definecolor{codepurple}{rgb}{0.58,0,0.82}
\definecolor{backcolour}{rgb}{1,1,1}
\lstdefinestyle{mystyle}{
  backgroundcolor=\color{backcolour}, commentstyle=\color{codegreen},
  keywordstyle=\color{codepurple},
  numberstyle=\tiny\color{codegray},
  stringstyle=\color{codepurple},
  basicstyle=\ttfamily\footnotesize,
  breakatwhitespace=false,         
  breaklines=true,                 
  captionpos=b,                    
  keepspaces=true,                 
  numbers=left,                    
  numbersep=5pt,                  
  showspaces=false,                
  showstringspaces=false,
  showtabs=false,                  
  tabsize=2
}
\usepackage{float}
\usepackage[frozencache,cachedir=.]{minted}
\usepackage{adjustbox}
\usepackage{tcolorbox}
\usepackage{amsfonts}
\usepackage{multirow}
\usepackage{verbatim}
\usepackage[font=small]{caption}
\usepackage{graphicx, subfig}
\usepackage{cleveref}
\usepackage{tabularx}
\usepackage{enumitem}
\usepackage{makecell}
\usepackage{siunitx}
\usepackage{xcolor}
\usepackage{threeparttable}

\newcolumntype{Y}{>{\centering\arraybackslash}X}
\usepackage{amsmath,bm}
\usepackage{microtype}
\usepackage{caption}
\usepackage{array}

\newcolumntype{C}[1]{>{\centering\arraybackslash}p{#1}}

\def\BibTeX{{\rm B\kern-.05em{\sc i\kern-.025em b}\kern-.08em
    T\kern-.1667em\lower.7ex\hbox{E}\kern-.125emX}}

\usepackage{enumitem}
\setlist[itemize]{noitemsep, topsep=0pt}

\icmltitlerunning{ChipNeMo: Domain-Adapted LLMs for Chip Design}

\makeatletter
\g@addto@macro\@verbatim\footnotesize
\makeatother
\begin{document}

\title{ChipNeMo: Domain-Adapted LLMs for Chip Design}

\twocolumn[
\icmltitle{ChipNeMo: Domain-Adapted LLMs for Chip Design}


\icmlsetsymbol{equal}{*}

\begin{icmlauthorlist}
\icmlauthor{Mingjie Liu}{equal,nvidia}
\icmlauthor{Teodor-Dumitru Ene}{equal,nvidia}
\icmlauthor{Robert Kirby}{equal,nvidia}
\icmlauthor{Chris Cheng}{equal,nvidia}
\icmlauthor{Nathaniel Pinckney}{equal,nvidia}
\icmlauthor{Rongjian Liang}{equal,nvidia}
\icmlauthor{Jonah Alben}{nvidia}
\icmlauthor{Himyanshu Anand}{nvidia}
\icmlauthor{Sanmitra Banerjee}{nvidia}
\icmlauthor{Ismet Bayraktaroglu}{nvidia}
\icmlauthor{Bonita Bhaskaran}{nvidia}
\icmlauthor{Bryan Catanzaro}{nvidia}
\icmlauthor{Arjun Chaudhuri}{nvidia}
\icmlauthor{Sharon Clay}{nvidia}
\icmlauthor{Bill Dally}{nvidia}
\icmlauthor{Laura Dang}{nvidia}
\icmlauthor{Parikshit Deshpande}{nvidia}
\icmlauthor{Siddhanth Dhodhi}{nvidia}
\icmlauthor{Sameer Halepete}{nvidia}
\icmlauthor{Eric Hill}{nvidia}
\icmlauthor{Jiashang Hu}{nvidia}
\icmlauthor{Sumit Jain}{nvidia}
\icmlauthor{Ankit Jindal}{nvidia}
\icmlauthor{Brucek Khailany}{nvidia}
\icmlauthor{George Kokai}{nvidia}
\icmlauthor{Kishor Kunal}{nvidia}
\icmlauthor{Xiaowei Li}{nvidia}
\icmlauthor{Charley Lind}{nvidia}
\icmlauthor{Hao Liu}{nvidia}
\icmlauthor{Stuart Oberman}{nvidia}
\icmlauthor{Sujeet Omar}{nvidia}
\icmlauthor{Ghasem Pasandi}{nvidia}
\icmlauthor{Sreedhar Pratty}{nvidia}
\icmlauthor{Jonathan Raiman}{nvidia}
\icmlauthor{Ambar Sarkar}{nvidia}
\icmlauthor{Zhengjiang Shao}{nvidia}
\icmlauthor{Hanfei Sun}{nvidia}
\icmlauthor{Pratik P Suthar}{nvidia}
\icmlauthor{Varun Tej}{nvidia}
\icmlauthor{Walker Turner}{nvidia}
\icmlauthor{Kaizhe Xu}{nvidia}
\icmlauthor{Haoxing Ren}{nvidia}
\end{icmlauthorlist}

\icmlaffiliation{nvidia}{NVIDIA}


\icmlkeywords{Large Language Models, Domain Adaptation, Chip Design, ICML}

\vskip 0.3in
]



\printAffiliationsAndNotice{\icmlEqualContribution} 


\begin{abstract}
ChipNeMo aims to explore the applications of large language models (LLMs) for industrial chip design.  Instead of directly deploying off-the-shelf commercial or open-source LLMs, we instead adopt the following domain adaptation techniques: domain-adaptive tokenization, domain-adaptive continued pretraining, model alignment with domain-specific instructions, and domain-adapted retrieval models. We evaluate these methods on three selected LLM applications for chip design: an engineering assistant chatbot,  EDA script generation, and bug summarization and analysis. 
Our evaluations demonstrate that domain-adaptive pretraining of language models, can lead to superior performance in domain related downstream tasks compared to their base LLaMA2 counterparts, without degradations in generic capabilities. 
In particular, our largest model, ChipNeMo-70B, outperforms the highly capable GPT-4 on two of our use cases, namely engineering assistant chatbot and EDA scripts generation, while exhibiting competitive performance on bug summarization and analysis. These results underscore the potential of domain-specific customization for enhancing the effectiveness of large language models in specialized applications.
\end{abstract}


\section{Introduction}
\label{sec:intro}


Over the last few decades, Electronic Design Automation (EDA) algorithms and tools have provided huge gains in chip design productivity.  Coupled with the exponential increases in transistor densities provided by Moore's law, EDA has enabled the development of feature-rich complex SoC designs with billions of transistors.  More recently, researchers have been exploring ways to apply AI to EDA algorithms and the chip design process to further improve chip design productivity~\cite{IEEEMicro2020,NVCell,prefixRL}. However, many time-consuming chip design tasks that involve interfacing with natural languages or programming languages still have not been automated.  The latest advancements in commercial (ChatGPT, Bard, etc.) and open-source (Vicuna \cite{vicuna2023}, LLaMA2 \cite{touvron2023llama}, etc.) large language models (LLM)  provide an unprecedented opportunity to help automate these language-related chip design tasks. Indeed, early academic research~\cite{benchmark,chipchat,chateda} has explored applications of LLMs for generating Register Transfer Level (RTL) code that can perform simple tasks in small design modules as well as generating scripts for EDA tools. 


We believe that LLMs have the potential to help chip design productivity by using generative AI to automate many language-related chip design tasks such as code generation, responses to engineering questions via a natural language interface, analysis and report generation, and bug triage. 
 In this study, we focus on three specific LLM applications: an \textbf{engineering assistant chatbot} for GPU ASIC and Architecture design engineers, which understands internal hardware designs and is capable of explaining complex design topics; \textbf{EDA scripts generation} for two domain specific tools based on Python and Tcl for VLSI timing analysis tasks specified in English; \textbf{bug summarization and analysis} as part of an internal bug and issue tracking system. 

Although general-purpose LLMs trained on vast amounts of internet data exhibit remarkable capabilities in generative AI tasks across diverse domains (as demonstrated in \cite{bubeck2023sparks}), recent work such as BloombergGPT \cite{wu2023bloomberggpt} and BioMedLLM \cite{biomed} demonstrate that domain-specific LLM models can outperform a general purpose model on domain-specific tasks.   In the hardware design domain,  \cite{benchmark, liu2023verilogeval} showed that open-source LLMs (CodeGen \cite{nijkamp2022codegen}) fine-tuned on additional Verilog data can outperform state-of-art OpenAI GPT-3.5 models. Customizing LLMs in this manner also avoids security risks associated with sending proprietary chip design data to third party LLMs via APIs. However, it would be prohibitively expensive to train domain-specific models for every domain from scratch, since this often requires millions of GPU training hours.  
To cost-effectively train domain-specific models, we instead propose to combine the following techniques: 
Domain-Adaptive Pre-Training (DAPT)~\cite{gururangan2020dont} of foundation models with domain-adapted tokenizers, model alignment using general and domain-specific instructions, and retrieval-augmented generation (RAG) \cite{lewis2021retrievalaugmented} with a trained domain-adapted retrieval model.


\begin{figure}[t]
    \centering
    \includegraphics[trim={2.2in 1.5in 2in 1.5in},clip, width=.48\textwidth]{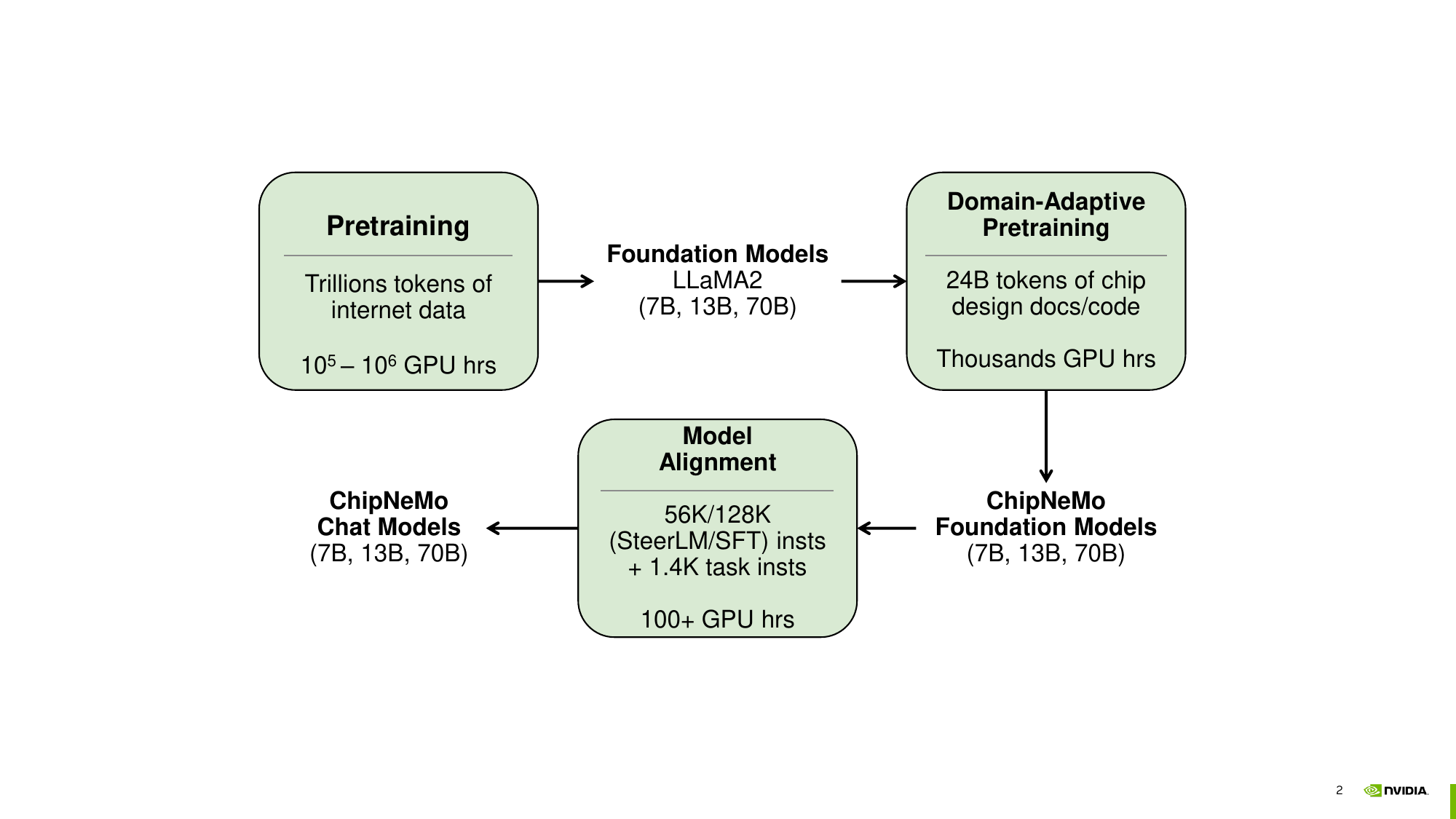}
    \caption{ChipNeMo Training Flow}
    \label{fig:training}
    \vspace{-0.2in}
\end{figure}

As shown in Figure \ref{fig:training}, our approach is to start with a base foundational model and apply DAPT followed by model alignment.  
DAPT, also known as continued pretraining with in-domain data, has been shown to be effective in areas such as biomedical and computer science publications, news, and reviews.  
In our case, we construct our domain-specific pre-training dataset from a collection of proprietary hardware-related code (e.g. software, RTL, verification testbenches, etc.) and natural language datasets (e.g. hardware specifications, documentation, etc.).  We clean up and preprocess the raw dataset, then continued-pretrain a foundation model with the domain-specific data. We call the resulting model a ChipNeMo foundation model. DAPT is done on a fraction of the tokens used in pre-training, and is much cheaper, only requiring roughly 1.5\% of the pretraining compute. 

LLM tokenizers convert text into sequences of tokens for training and inference.  A domain-adapted tokenizer improves the tokenization efficiency by tailoring rules and patterns for domain-specific terms such as keywords commonly found in RTL. For DAPT, we cannot retrain a new domain-specific tokenizer from scratch, since it would make the foundation model invalid. Instead of restricting ChipNeMo to the pre-trained general-purpose tokenizer used by the foundation model, we instead adapt the pre-trained tokenizer to our chip design dataset, only adding new tokens for domain-specific terms.

ChipNeMo foundation models are completion models which require model alignment to adapt to tasks such as chat. We use largely publicly available general-purpose chat instruction datasets for multi-turn chat together with a small amount of domain-specific instruction datasets to perform alignment on the ChipNeMo foundation model,  which produces the ChipNeMo chat model. 
We observe that alignment with a general purpose chat instruction dataset is adequate to align the ChipNeMo foundation models with queries in the chip design domain. We also added a small amount of task-specific instruction data, which further improves the alignment. We trained multiple ChipNeMo foundation and chat models based on variants of LLaMA2 models used as the base foundation model. 


To improve performance on the engineering assistant chatbot application, we also leverage Retrieval Augmented Generation (RAG).  RAG is an \textit{open-book} approach for giving LLMs precise context for user queries. It retrieves relevant in-domain knowledge from its data store to augment the response generation given a user query. This method shows significant improvement in grounding the model to the context of a particular question. Crucially we observed significant improvements in retrieval hit rate when finetuning a pretrained retrieval model with domain data. This led to even further improvements in model quality.





Our results show that domain-adaptive pretraining was the primary technique driving enhanced performance in domain-specific tasks. We highlight the following contributions and findings for adapting LLMs to the chip design domain:

\begin{itemize}
    \item We demonstrate domain-adapted LLM effectiveness on three use-cases: an engineering assistant chatbot, EDA tool script generation, and bug summarization and analysis. We achieve a score of 6.0 on a 7 point Likert scale for engineering assistant chatbot based on expert evaluations, more than 70\% correctness on the generation of simple EDA scripts, and expert evaluation ratings above 5 on a 7 point scale for summarizations and assignment identification tasks.
   
    
    \item Domain-adapted ChipNeMo models dramatically outperforms all vanilla LLMs evaluated on both multiple-choice domain-specific AutoEval benchmarks and human evaluations for applications. 

    \item Using the SteerLM alignment method~\cite{dong2023steerlm} over traditional SFT improves human evaluation scores for the engineering assistant chatbot by 0.62 points on a 7 point Likert scale.
    
    \item SFT on an additional $1.4K$ domain-specific instructions significantly improves the model's proficiency at generating correct EDA tool scripts by $18\%$.
    
     
    \item Domain-adaptive tokenization reduce domain data token count by up to $3.3\%$ without hurting effectiveness on applications.
    \item Fine-tuning our ChipNeMo retrieval model with domain-specific data improves the retriever hit rate by 30\% over a pre-trained state-of-the-art retriever, in turn improving overall quality of RAG responses.
\end{itemize}

The paper is organized as follows. 
Section \ref{sec:train} outlines domain adaptation and training methods used including the adapted tokenizer, DAPT, model alignment, and RAG. 
Section~\ref{sec:evaluation} describes the experimental results including human evaluations for each application. 
Section \ref{sec:related} describes relevant LLM methods and other work targeting LLMs for chip design.  
Finally, detailed results along with additional model training details and examples of text generated by the application use-cases are illustrated in the Appendix.






\vspace{3mm}


\section{ChipNeMo Domain Adaptation Methods}
\label{sec:train}

ChipNeMo implements multiple domain adaptation techniques to adapt LLMs to the chip design domain. These techniques include domain-adaptive tokenization for chip design data, domain adaptive pretraining with large corpus of domain data, model alignment to domain specific tasks, and retrieval-augmented generation with a fine-tuned retrieval model. 
We illustrate the details of each technique in this section.

\subsection{Domain-Adaptive Tokenization}
\label{sec:tokenizer}
When adapting a pre-trained tokenizer, the main goals are to improve tokenization efficiency on domain-specific data, maintain language model performance on general datasets, and minimize the effort for retraining/fine-tuning. To achieve this, we developed the following approach:
\begin{enumerate}
    \item Train a tokenizer from scratch using domain-specific data.
    \item From the vocabulary of the new tokenizer, identifying tokens that are absent in the general-purpose tokenizer and are rarely found in general-purpose datasets.
    \item Expand the general-purpose tokenizer with the newly identified tokens at Step~2.
    \item Initialize model embeddings of the new tokens by utilizing the general-purpose tokenizer.

    
\end{enumerate}
    
Specifically for Step 4, when a new token is encountered, it is first re-tokenized using the original pretrained general-purpose tokenizer. The LLM's token embedding for the new token is determined by averaging the embeddings of the tokens generated by the general-purpose tokenizer~\cite{koto-etal-2021-indobertweet}. The LLM's final output layer weights for the new tokens are initialized to zero.

Step~2 helps maintain the performance of the pre-trained LLM on general datasets by selectively introducing new tokens that are infrequently encountered in general-purpose datasets. Step~4 reduces the effort required for retraining or finetuning the LLM via initialization of the embeddings of new tokens guided by the general-purpose tokenizer.

\subsection{Domain Adaptive Pretraining} 

In our study, we apply DAPT on pretrained foundation base models: LLaMA2 7B/13B/70B. Each DAPT model is initialized using the weights of their corresponding pretrained foundational base models. We name our domain-adapted models \textbf{ChipNeMo}. We employ tokenizer augmentation as depicted in Section~\ref{sec:tokenizer} and initialize embedding weight accordingly~\cite{koto-etal-2021-indobertweet}. We conduct further pretraining on domain-specific data by employing the standard autoregressive language modeling objective. All model training procedures are conducted using the NVIDIA NeMo framework~\cite{kuchaiev2019nemo}, incorporating techniques such as tensor parallelism~\cite{shoeybi2019megatron} and flash attention~\cite{dao2022flashattention} for enhanced efficiency. 

Our models undergo a consistent training regimen with similar configurations. A small learning rate of $5 \cdot 10^{-6}$ is employed, and training is facilitated using the Adam optimizer, without the use of learning rate schedulers. The global batch size is set at 256, and a context window of 4096 tokens is applied, resulting in an effective batch size of 1M tokens. 
The total number of training steps is set to 23,200, equating to roughly 1 epoch of the data blend.

\begin{figure}[ht]
\vspace{-0.15in}
    \centering
    \includegraphics[width=.42\textwidth]{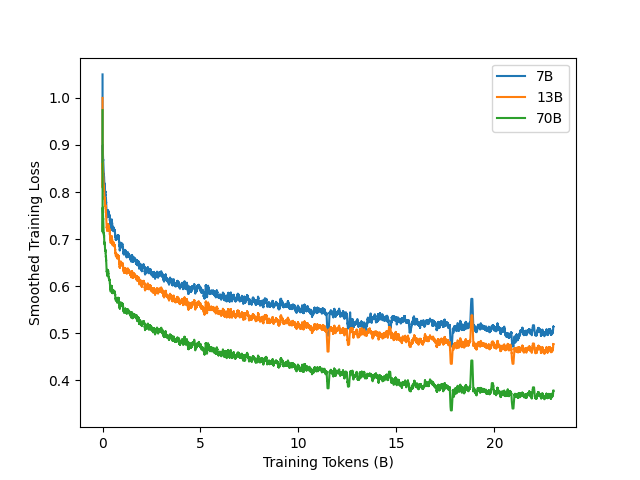}
    \caption{Smoothed Training Loss for ChipNeMo with Tokenizer Augmentation.}
    \label{fig:loss_llama2}
    \vspace{-0.1in}
\end{figure}

Figure~\ref{fig:loss_llama2} illustrates the training loss of ChipNeMo under the specified hyperparameters. We do observe spikes in the training loss. In contrast to the hypothesis in~\cite{chowdhery2022palm}, we postulate that in our scenario, these spikes can be attributed to ``bad data" since these irregularities seem to consistently occur in similar training steps for the same model, even across different model sizes. We chose not to address this issue, as these anomalies did not appear to significantly impede subsequent training steps (with no noticeable degradation in validation loss), possibly due to our application of a low learning rate.

We refer readers to Appendix for details on the training data collection process~\ref{app:data}, training data blend~\ref{sec:appendix:data}, and implementation details and ablation studies on domain-adaptive pretraining~\ref{sec:training}.

\subsection{Model Alignment}
\label{sec:SFT}
After DAPT, we perform model alignment. We specifically leverage two alignment techniques: supervised fine-tuning (SFT) and SteerLM~\cite{dong2023steerlm}. We adopt the identical hyperparameter training configuration as DAPT for all models, with the exception of using a reduced global batch size of 128. 
We employ an autoregressive optimization objective, implementing a strategy where losses associated with tokens originating from the system and user prompts are masked~\cite{touvron2023llama}. This approach ensures that during backpropagation, our focus is exclusively directed towards the optimization of answer tokens. 

We combined our domain alignment dataset, consisting of approximately 1.4k samples, with larger general chat datasets. For SFT, we blended the domain instructional data with 128k commercial-viable chat data and then performed fine-tuning for a single epoch after random shuffling. We conducted experiments involving augmentation of the domain-specific SFT dataset for more than one epoch. However, it became apparent that the model rapidly exhibited signs of overfitting when presented with in-domain questions, often repeating irrelevant answers from the domain SFT dataset. For SteerLM, we closely followed the steps outlined in~\cite{wang2023helpsteer}. We first trained an attribute model instantiated with LLaMA2-13B model on the HelpSteer and OASST datasets. We then used the attribute model to label all attributes for OASST data and our domain instructional data. Finally, we conducted attribute-conditioned fine-tuning and also masked the attribute labels and trained on ChipNeMo models for 2 epochs. We refer readers to Appendix~\ref{sec:appendix:data:sft} for details on the alignment datasets and~\ref{sec:appendix:training:sft} on implementations details.

We also experimented with DAPT directly on a chat aligned model, such as the LLaMA2-Chat model. We found that DAPT significantly degraded the model's alignment, making the resulting model useless for downstream tasks.

\subsection{Domain-Adapted Retrieval Model}
\label{sec:rag}
\begin{figure}[ht]
\vspace{-0.1in}
    \centering
    \includegraphics[width=.48\textwidth]{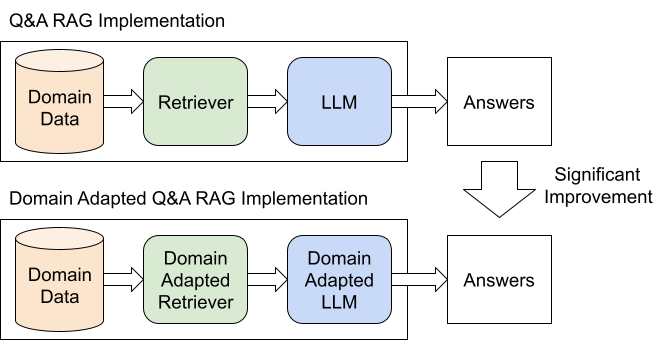}
    \caption{RAG Implementation Variations}
    \label{fig:rag_diagram}
    \vspace{-0.1in}
\end{figure}

It is well known that LLMs can generate inaccurate text, so-called \textit{hallucination} \cite{hallucination}. Although the phenomenon is not completely understood, we still must mitigate \textit{hallucinations} since they are particularly problematic in an engineering assistant chatbot context, where accuracy is critical. Our proposal is to leverage the retrieval augmented generation (RAG) method. RAG tries to retrieve relevant passages from a database to be included in the prompt together with the question, which grounds the LLM to produce more accurate answers. We find that using a domain adapted language model for RAG significantly improves answer quality on our domain specific questions. Also, we find that fine-tuning an off-the-shelf unsupervised pre-trained dense retrieval model with a modest amount of domain specific training data significantly improves retrieval accuracy. Our domain-adapted RAG implementation diagram is illustrated on Figure \ref{fig:rag_diagram}.

We created our domain adapted retrieval model by fine-tuning the \textit{e5\_small\_unsupervised} model~\cite{e5} with 3000 domain specific auto-generated samples using the Tevatron framework \cite{gao2022tevatron}. We refer readers to the details on the sample generation and training process in Appendix \ref{sec:retrieval_model_training}.

Even with the significant gains that come with fine-tuning a retrieval model, the fact remains that retrieval still struggles with queries that do not map directly to passages in the document corpus or require more context not present in the passage. Unfortunately, these queries are also more representative of queries that will be asked by engineers in real situations. Combining retrieval with a domain adapted language model is one way to address this issue.

\vspace{3mm}


\section{Evaluations}
\label{sec:evaluation}
We evaluate our training methodology and application performance in this section. We study our 7B, 13B, and 70B models in the training methodology evaluation, and only our ChipNeMo-70B  model using SteerLM for model alignment in the application performance evaluation. For comparison, we also evaluate two baseline chat models: LLaMA2-70B-Chat and GPT-4. LLaMA2-70B-Chat is the publicly released LLaMA2-Chat model trained with RLHF and is considered to be the state-of-the-art open-source chat model, while GPT-4 is considered to be the state-of-the-art proprietary chat model.

\begin{figure}[ht]
    \centering
    \includegraphics[width=.4\textwidth]{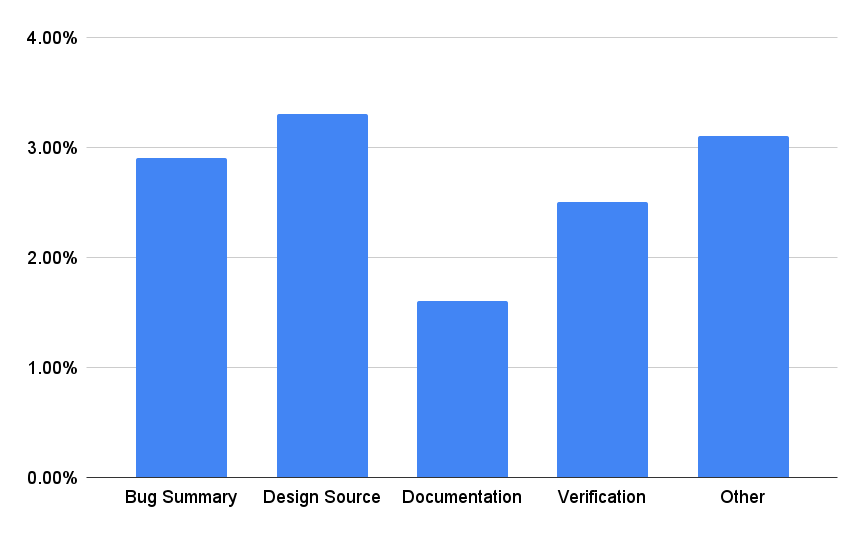}
    \caption{Domain-Adapted ChipNeMo Tokenizer Improvements.}
    \label{fig:tokenizer}
    \vspace{-0.1in}
\end{figure}

\subsection{Domain-Adaptive Tokenization}
We adapt the LLaMA2 tokenizer (containing 32K tokens) to chip design datasets using the previously outlined four-step process. Approximately 9K new tokens are added to the LLaMA2 tokenizer. The adapted tokenizers can improve tokenization efficiency by 1.6\% to 3.3\% across various chip design datasets as shown in Figure~\ref{fig:tokenizer}. We observe no obvious changes to tokenizer efficiency on public data. Importantly, we have not observed significant decline in the LLM's accuracy on public benchmarks when using the domain-adapted tokenizers even prior to DAPT.

\subsection{Domain Adaptive Pretraining}
\begin{figure}[htbp]
  \centering
  \vspace{-0.1in}
  \includegraphics[width=0.48\textwidth]{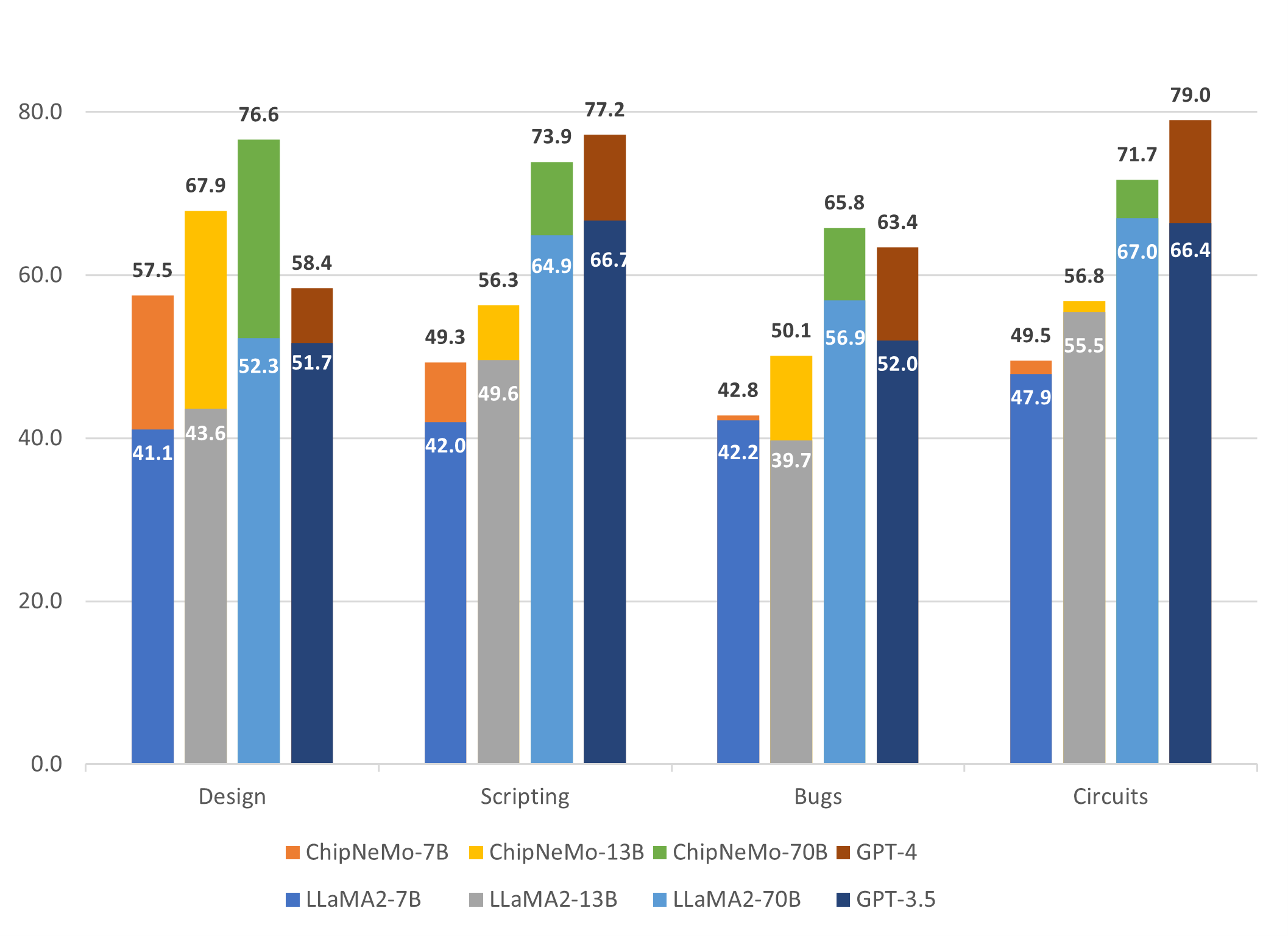}
  \caption{Chip Domain Benchmark Result for ChipNeMo.}
  \label{fig:chipllama2_benchmark}
\end{figure}

Figure~\ref{fig:chipllama2_benchmark} presents the outcomes for ChipNeMo models on the AutoEval benchmark for chip design domain (detailed in Appendix~\ref{sec:appendix:evaluation}). Results on open domain academic benchmark results are presented in Appendix~\ref{sec:training}. Our research findings can be summarized as follows:
\begin{itemize}
    \item DAPT exerts a substantial positive impact on tasks within the domain itself. This effect is manifested in significant improvements in internal design knowledge as well as general circuit design knowledge.
    \item DAPT models exhibit a slight degradation in performance on open-domain academic benchmarks.
    \item The use of larger and more performant foundational models yields better zero-shot results on domain-specific tasks. Furthermore, the employment of superior base models results in enhanced domain models post-DAPT, leading to heightened performance on in-domain tasks.
    \item Improvements attributed to DAPT with in-domain tasks exhibit a positive correlation with model size, with larger models demonstrating more pronounced enhancements in domain-specific task performance.
\end{itemize}

\subsection{Training Ablation Studies}
For our ablation studies, we conducted multiple rounds of domain adaptive pre-training. We provide brief summaries and refer to the Appendix~\ref{sec:training} for details.

The differences between training with the augmented tokenizer and the original tokenizer appeared to be negligible. We thus primarily attribute the accuracy degradation on open-domain academic benchmarks to domain data. Moreover, the removal of the public dataset only slightly regressed on most tasks including academic benchmarks. 

In our exploration, we experimented with employing a larger learning rate, as in CodeLLaMA~\cite{rozière2023code}. We observed large spikes in training loss at the initial training steps. Although this approach eventually led to improved training and validation loss, we noted substantial degradations across all domain-specific and academic benchmarks, except on coding. We hypothesize that a smaller learning rate played a dual role, facilitating the distillation of domain knowledge through DAPT while maintaining a balance that did not veer too far from the base model, thus preserving general natural language capabilities.

We also explored the application of Parameter Efficient Fine-Tuning (PEFT) in the context of Domain-Adaptive Pre-training (DAPT). In this pursuit, we conducted two experiments involving the incorporation of LoRA adapters~\cite{LoRA}, introducing additional parameters of 26.4 million (small) and 211.2 million (large) respectively. In both instances, our findings revealed a significant accuracy gap on in-domain tasks when compared to the full-parameter DAPT approach. Furthermore, when contrasting the outcomes between small and large PEFT models, we observed a marginal enhancement on in-domain task accuracy, with large adapter exhibiting a slight improvement. 


\subsection{Training Cost}

All models have undergone training using 128 A100 GPUs. We estimate the costs associated with domain adaptive pretraining for ChipNeMo as illustrated in Table~\ref{tab:training_costs}. It is worth noting that DAPT accounts for less than 1.5\% of the overall cost of pretraining a foundational model from scratch.

\begin{table}[ht]
    \small
    \centering
    \begin{tabular}{|l|c|c|c|}
        \hline
        Model Size & Pretraining &   DAPT & SFT \\
        \hline
        7B & 184,320 & 2,620 & 90\\
        13B & 368,640 & 4,940 & 160 \\
        70B & 1,720,320 & 20,500  & 840\\
        \hline
    \end{tabular}
    \caption{Training cost of LLaMA2 models in A100 GPU hours. Pretraining cost from~\cite{touvron2023llama}.}
    \label{tab:training_costs}
\end{table}

\subsection{RAG and Engineering Assistant Chatbot}
\label{sec:rag_chat_eval}
We created a benchmark to evaluate the performance of design chat assistance, which uses the RAG method. This benchmark includes 88 questions in three categories: 
architecture/design/verification specifications (Specs), testbench regression documentation (Testbench), and build infrastructure documentation (Build). For each question, we specify the golden answer as well as the paragraphs in the design document that contains the relevant knowledge for the answer. These questions are created by designers manually based on a set of design documents as the data store for retrieval. It includes about 1.8K documents, which were segmented into 67K passages, each about 512 characters.

First, we compare our domain adapted retrieval model with Sentence Transformer \cite{reimers-2019-sentence-bert} and \textit{e5\_small\_unsupervised} \cite{e5} on each category. Each model fetches its top 8 passages from the data store.

\begin{figure}[ht]
    \centering
    \includegraphics[ width=.48\textwidth]{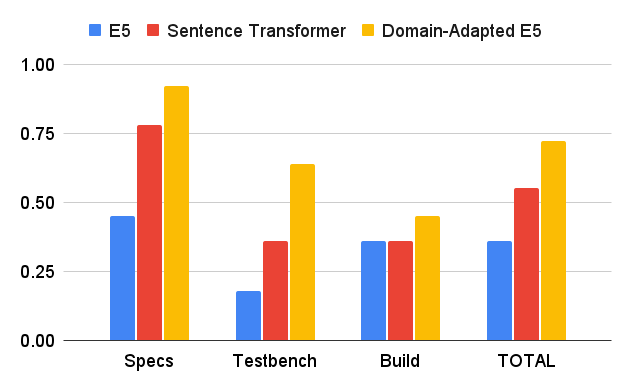}
    \caption{Retrieval Model Accuracy Comparison}
    \label{fig:retieval_acc}
    \vspace{-0.1in}
\end{figure}

As shown in Figure \ref{fig:retieval_acc}, our domain-adapted model performed 2x better than the original \textit{e5\_small\_unsupervised} model and 30\% better than sentence transformer.

The queries in the Specs category are derived directly from passages in the documents, so their answers are often nicely contained in a concise passage and clearly address the query. On the other hand, the queries of the Testbench and Build categories are not directly derived from passages, so their answers were often not as apparent in the fetched passages and required more context (see Appendix~\ref{sec:retrieval_model_training} for detailed examples). This significantly contributes to the difference in retrieval quality between the categories.

\begin{figure}[ht]
    \centering
    \includegraphics[width=.48\textwidth]{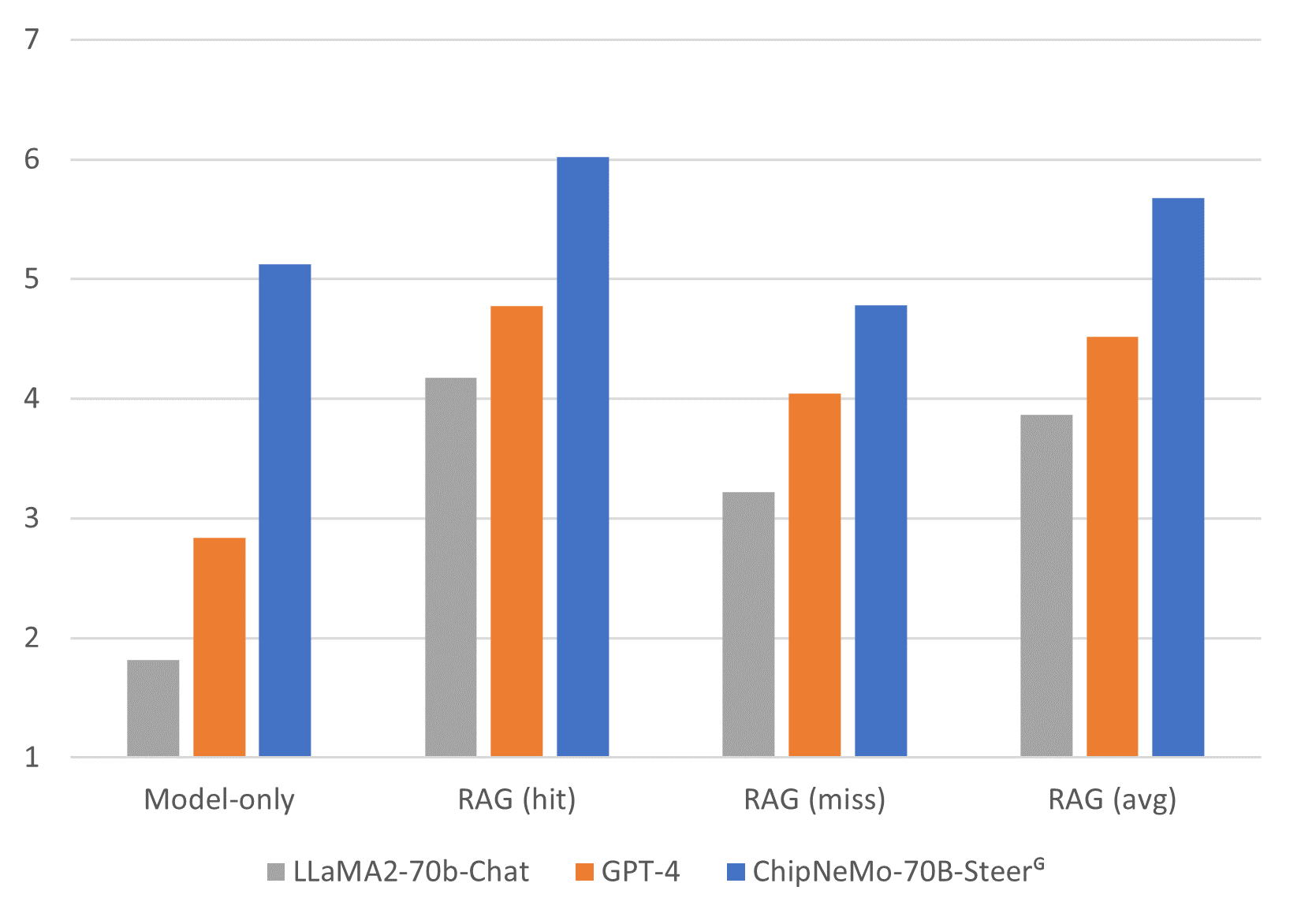}
    \caption{Human Evaluation of Different Models. Model Only represents results without RAG. RAG (hit)/(miss) only include questions whose retrieved passages hit/miss their ideal context, RAG (avg) includes all questions. 7 point Likert scale.}
    \label{fig:human_eval}
\end{figure}

We conducted evaluation of multiple ChipNeMo models and LLaMA2 models with and without RAG. The results were then scored by human evaluators on a 7 point Likert scale and shown in Figure \ref{fig:human_eval}.
We highlight the following:
\begin{itemize}
    
    
    \item ChipNeMo-70B-Steer outperforms GPT-4 in all categories, including both RAG misses and hits.
    \item ChipNeMo-70B-Steer outperforms similar sized LLaMA2-70b-Chat in model-only and RAG evaluations by 3.31 and 1.81, respectively.
     

\end{itemize}

Our results indicate that RAG significantly boosts human scores. RAG improves ChipNeMo-70B-Steer, GPT-4, and LLaMA2-70b-Chat by 0.56, 1.68, and 2.05, respectively. Even when RAG misses, scores are generally higher than without using retrieval. The inclusion of relevant in-domain context still led to improved performance, as retrieval is not a strictly binary outcome. Furthermore, while ChipNeMo-70B-SFT outperforms GPT4 by a large margin through traditional supervised fine-tuning, applying SteerLM methods~\cite{wang2023helpsteer} leads to further elevated chatbot ratings. We refer readers to the complete evaluation results in Appendix~\ref{app:chat}.

\subsection{EDA Script Generation}
\label{sec:vivid}

In order to evaluate our model on the EDA script generation task, we created two different types of benchmarks. The first is a set of ``Easy'' and ``Medium'' difficulty tasks (1-4 line solutions) that can be evaluated without human intervention by comparing with a golden response or comparing the generated output after code execution. 
The second set of tasks ``Hard'' come from real use case scenarios that our engineers chose. These tasks are much harder requiring multiple API calls and understanding relationship between different VLSI objects. Because these are hard to evaluate in an automated way (with current model performance), we had human engineers judge the correctness between 0-10.
We evaluate the model on two tools, one is a fully in-house Python based tool and the other is a Tcl based EDA tool with limited public data.
The size of these benchmarks are described in Table \ref{tab:vividpt_eval_description}. Work is ongoing to both increase the size and scope for these benchmarks to allow us to further assess and improve these models. 

\begin{table}[ht]
\small
\centering
\begin{tabular}{|c|c|}
   \hline
   Evaluation Benchmark Name  & Size \\
   \hline
   Python Tool - Automatic (Easy)  & 146 \\
   Python Tool - Automatic (Medium) & 28 \\
   Python Tool -  Human (Hard) & 25 \\
   Tcl Tool - Automatic (Easy)  & 708 \\
   Tcl Tool - Automatic (Medium) & 27 \\
   Tcl Tool -  Human (Hard) & 25 \\
   \hline
\end{tabular}
\caption{EDA Script Generation Evaluation Benchmarks}
\label{tab:vividpt_eval_description}
\end{table}

\begin{figure}[ht]
    \centering
    \includegraphics[trim={0in 0in 0in 0in},clip, width=.45\textwidth]{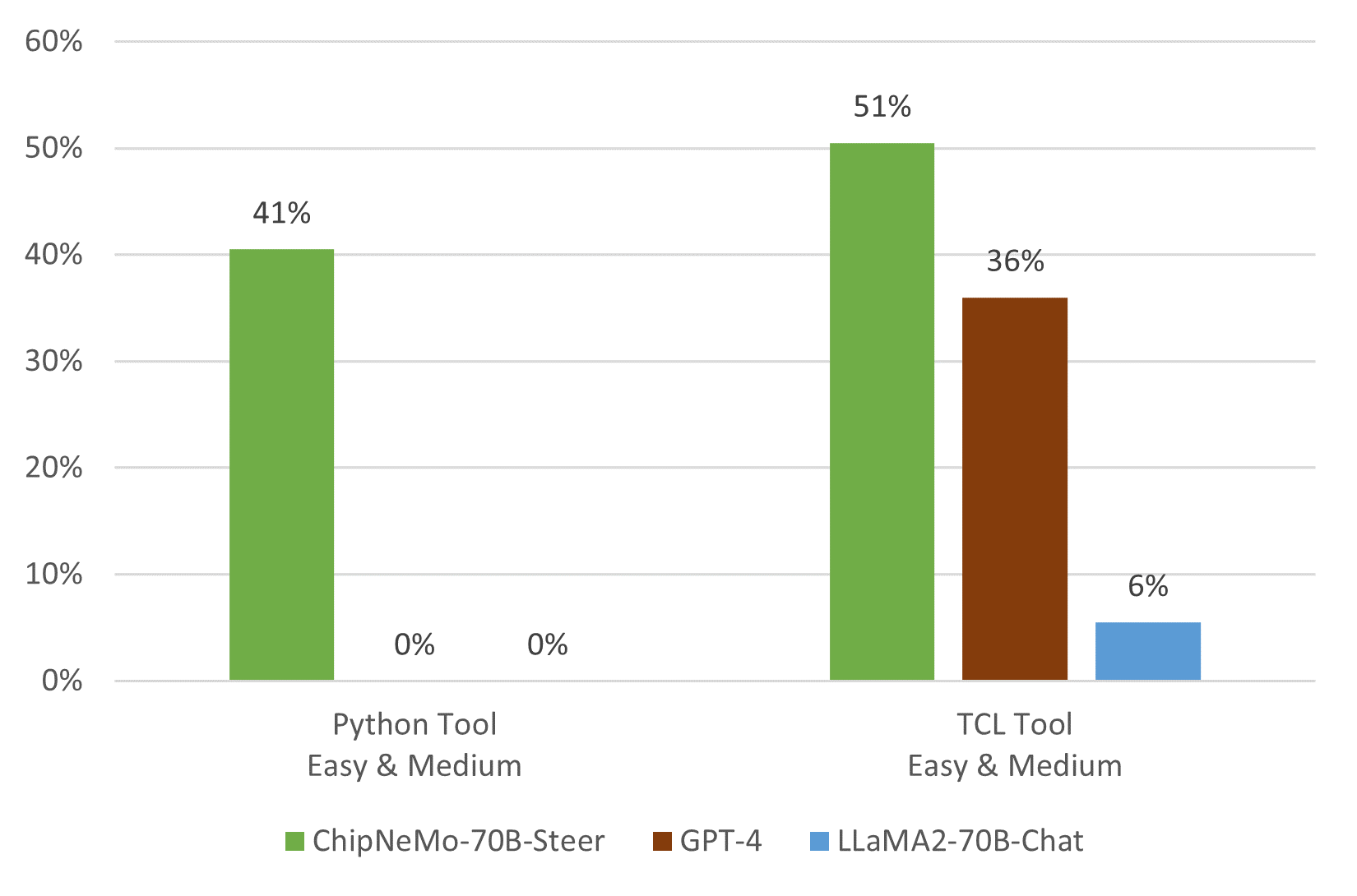}
    \caption{EDA Script Generation Evaluation Results, Pass@5}
    \label{fig:eda_results1}
\end{figure}
\begin{figure}[ht]
    \centering
    \includegraphics[trim={0in 0in 0in 0in},clip, width=.45\textwidth]{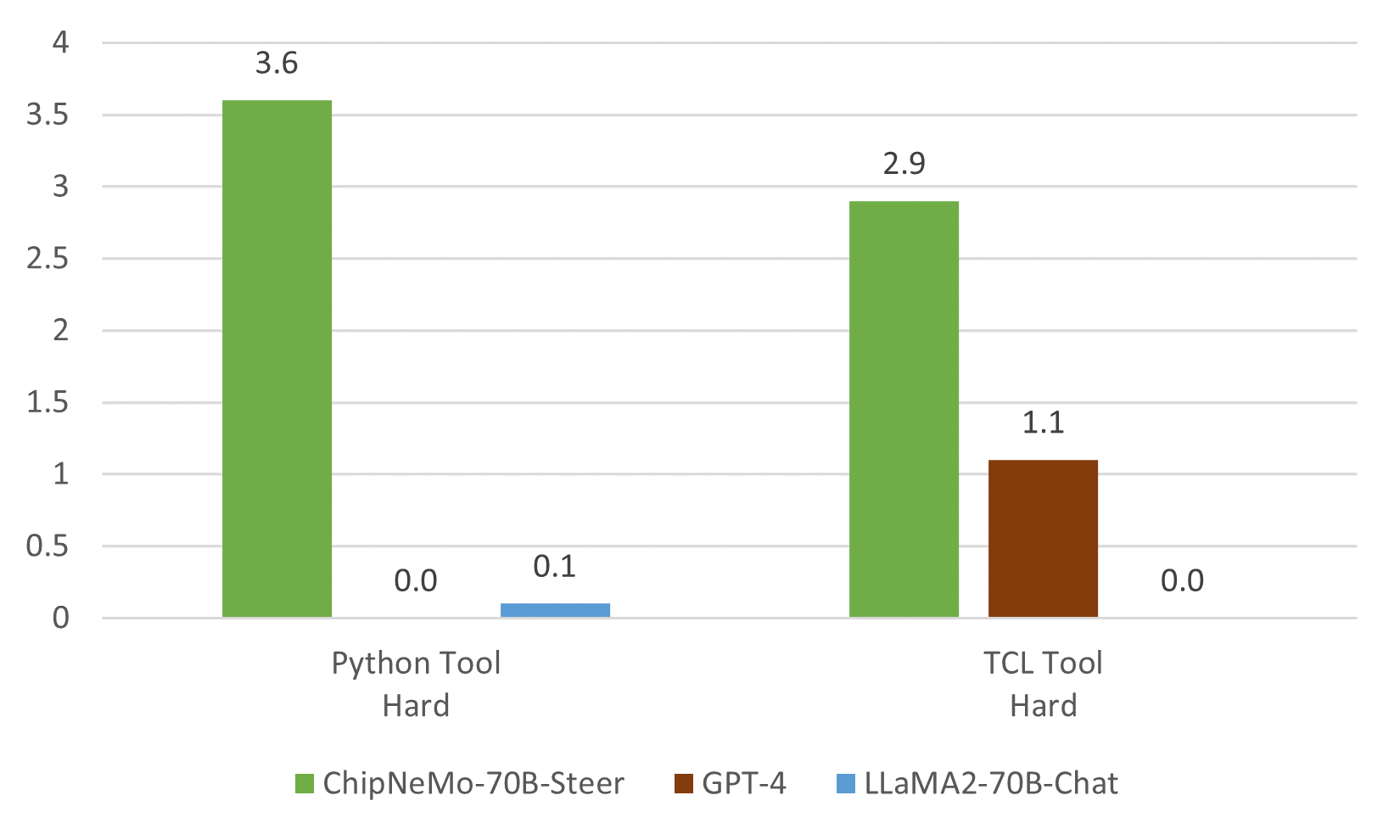}
    \caption{EDA Script Generation Evaluation Results, Single Generation (temperature=0), Human Evaluated 0-10 Point Scale.}
    \label{fig:eda_results2}
    \vspace{-0.1in}
\end{figure}

The comparative performance of our models on these evaluations are shown in Figures~\ref{fig:eda_results1} and~\ref{fig:eda_results2}.
Figure~\ref{fig:eda_results1} shows the results on automated ``easy'' and ``medium'' benchmarks where we check for fully accurate code. For ``Hard'' benchmarks in Figure~\ref{fig:eda_results2} we check for partial correctness of the code, which is evaluated by a human user on a 0-10 scale. ChipNeMo-70B-Steer performs significantly better than off-the-shelf GPT-4 and LLaMA2-70B-Chat model. 


As seen in Figure~\ref{fig:eda_results1}, models like GPT-4 and LLaMA2-70B-Chat have close to zero accuracy for the Python tool where the domain knowledge related to APIs of the tool are necessary. This shows the importance of DAPT. Without DAPT, the model had little to no understanding of the underlying APIs and performed poorly on both automatic and human evaluated benchmarks. Our aligned model further improved the results of DAPT because our domain instructional data helps guide the model to present the final script in the most useful manner. An ablation study on inclusion of domain instructional data for model alignment and the application of retrieval is provided in Appendix \ref{sec:appendix:additional_eval}.


Our non-domain models performed better on our Tcl tool than the Python tool, but the trend for our domain trained model was the opposite. We suspect this was due to the proprietary nature of our Python tool. It was difficult for general LLMs to perform well on our Python tool benchmark without knowledge of the APIs. Since ChipNeMo is trained with domain data, the inherent python coding ability of the base model allows ChipNeMo-70B-Steer to perform better. This again highlights the importance of DAPT for low-volume or proprietary programming languages.


\begin{figure}[ht]
    \centering
    \includegraphics[trim={0in 0in 0in 0in},clip, width=.45\textwidth]{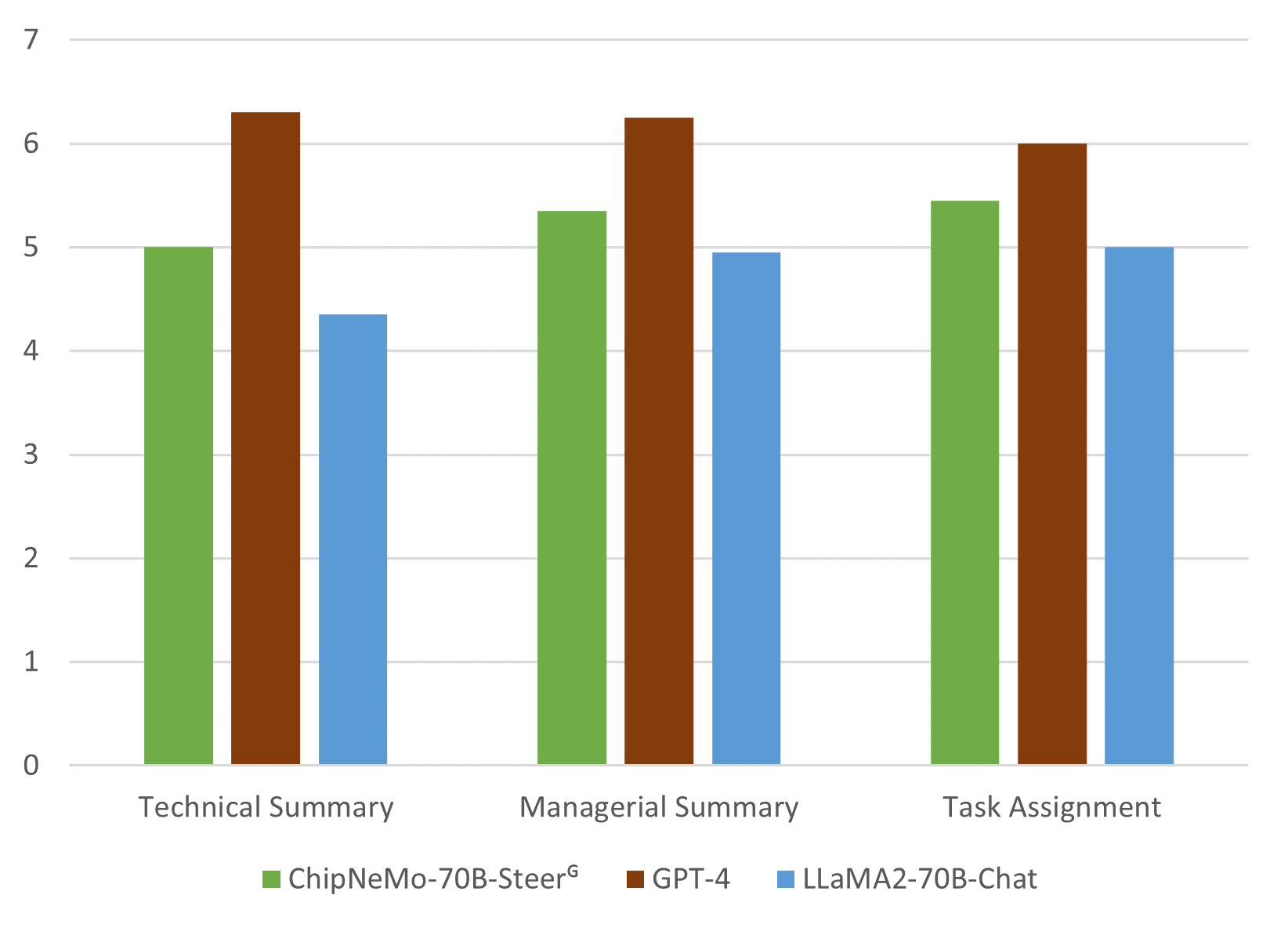}
    \caption{Bug Summarization and Analysis Evaluation Results, 7 point Likert scale.}
    \label{fig:nvbugs_results}
    \vspace{-0.1in}
\end{figure}

\subsection{Bug Summarization and Analysis}
\label{sec:nvbugs}

To evaluate our models on bug summarization and analysis we have a hold out set of 30 bugs which are ideal candidates for summarization. This includes having a long comment history or other data which makes the bugs hard for a human to quickly summarize. As described in Appendix~\ref{subsubsec:nvbugs} the long length of each individual bug requires the LLM to perform hierarchical summarization.

We study three separate sub-tasks: summarization focused on technical details, summarization focused on managerial details, and a post-summarization recommendation of task assignnment. Participants are tasked with rating the model's performance on a 7-point Likert scale for each of these three assignments. The results can be found in Figure~\ref{fig:nvbugs_results}. Although the GPT-4 model excels in all three tasks, outperforming both our ChipNeMo-70B-Steer model and the LLaMA2-70B-Chat model, ChipNeMo-70B-Steer does exhibit enhancements compared to the off-the-shelf LLaMA model of equivalent size. We attribute the comparatively lower improvements in summarization tasks resulting from our domain-adaptation to the limited necessity for domain-specific knowledge in summarization compared to other use-cases.

\vspace{3mm}


\section{Related Works} 
\label{sec:related}

Many domains have a significant amount of proprietary data which can be used to train a domain-specific LLM.  One approach is to train a domain specific foundation model from scratch, e.g., BloombergGPT\cite{wu2023bloomberggpt} for finance, BioMedLLM\cite{biomed} for biomed, and Galactica\cite{taylor2022galactica} for science. These models were usually trained on more than 100B tokens of raw domain data. The second approach is domain-adaptive pretraining (DAPT) \cite{gururangan2020dont} which continues to train a pretrained foundation model on additional raw domain data. It shows slight performance boost on domain-specific tasks in domains such as biomedical, computer science publications, news, and reviews. In one example, \cite{lewkowycz2022solving} continued-pretrained a foundation model on technical content datasets and achieved state-of-the-art performance on many quantitative reasoning tasks.

Retrieval Augmented Generation (RAG) helps ground the LLM to generate accurate information and to extract up-to-date information to improve knowledge-intensive NLP tasks \cite{RAG}. It is observed that smaller models with RAG can outperform larger models without RAG \cite{RETRO}. Retrieval methods include sparse retrieval methods such as TF-IDF or BM25\cite{BM25}, which analyze word statistic information and find matching documents with a high dimensional sparse vector. Dense retrieval methods such as \cite{DPR, contriever} find matching documents on an embedding space generated by a retrieval model pretrained on a large corpus with or without fine-tuning on a retrieval dataset. The retrieval model can be trained standalone \cite{DPR, contriever, REPLUG} or jointly with language models \cite{ATLAS, RETRO}. In addition, it has been shown that off-the-shelf general purpose retrievers can improve a baseline language model significantly without further fine-tuning \cite{incontext-RAG}. RAG is also proposed to perform code generation tasks \cite{docprompting} by retrieving from coding documents. 

Foundation models are completion models, which have limited chat and instruction following capabilities. Therefore,  a model alignment process is applied to the foundation models to train a corresponding chat model. Instruction fine-tuning \cite{wei2022finetuned} and reinforcement learning from human feedback (RLHF) \cite{ouyang2022training} are two common model alignment techniques. Instruction fine-tuning further trains a foundation model using instructions datasets. 
RLHF leverages human feedback to label a dataset to train a reward model and applies reinforcement learning to further improve models given the trained reward model. RLHF is usually more complex and resource hungry than instruction fine-tuning. Therefore, recent studies also propose to reduce this overhead with simpler methods such as DPO \cite{rafailov2023direct} and SteerLM \cite{dong2023steerlm}.

Researchers have started to apply LLM to chip design problems. Early works such as Dave \cite{dave} first explored the possibility of generating Verilog from English with a language model (GPT-2). Following that work, \cite{benchmark} showed that fine-tuned open-source LLMs (CodeGen) on Verilog datasets collected from GitHub and Verilog textbooks outperformed state-of-the-art OpenAI models such as \textit{code-davinci-002} on 17 Verilog questions. \cite{liu2023verilogeval} proposed a benchmark with more than 150 problems and demonstrated that the Verilog code generation capability of pretrained language models could be improved with supervised fine-tuning by bootstrapping with LLM generated synthetic problem-code pairs. Chip-Chat \cite{chipchat} experimented with conversational flows to design and verify a 8-bit accumulator-based microprocessor with GPT-4 and GPT-3.5. Their findings showed that although GPT-4 produced relatively high-quality codes, it still does not perform well enough at understanding and fixing the errors. ChipEDA \cite{chateda} proposed to use LLMs to generate EDA tools scripts. It also demonstrated that fine-tuned LLaMA2 70B model outperforms GPT-4 model on this task. 
\vspace{3mm}

\section{Conclusions} 
\label{sec:conclusion}
We explored domain-adapted approaches to improve LLM performance for industrial chip design tasks. Our results show that domain-adaptive pretrained models, such as the 7B, 13B, and 70B variants of ChipNeMo, achieve similar or better results than their base LLaMA2 models with only 1.5\% additional pretraining compute cost. Our largest trained model, ChipNeMo-70B, also surpasses the much more powerful GPT-4 on two of our use cases, engineering assistant chatbot and EDA scripts generation, while showing competitive performance on bug summarization and analysis.
Our future work will focus on further improving ChipNeMo models and methods for production use. 
\vspace{3mm}

\newpage
\bibliography{ref}

\begin{thebibliography}{56}
\providecommand{\natexlab}[1]{#1}
\providecommand{\url}[1]{\texttt{#1}}
\expandafter\ifx\csname urlstyle\endcsname\relax
  \providecommand{\doi}[1]{doi: #1}\else
  \providecommand{\doi}{doi: \begingroup \urlstyle{rm}\Url}\fi

\bibitem[Blocklove et~al.(2023)Blocklove, Garg, Karri, and Pearce]{chipchat}
Blocklove, J., Garg, S., Karri, R., and Pearce, H.
\newblock Chip-chat: Challenges and opportunities in conversational hardware design, 2023.

\bibitem[Borgeaud et~al.(2022)Borgeaud, Mensch, Hoffmann, Cai, Rutherford, Millican, van~den Driessche, Lespiau, Damoc, Clark, de~Las~Casas, Guy, Menick, Ring, Hennigan, Huang, Maggiore, Jones, Cassirer, Brock, Paganini, Irving, Vinyals, Osindero, Simonyan, Rae, Elsen, and Sifre]{RETRO}
Borgeaud, S., Mensch, A., Hoffmann, J., Cai, T., Rutherford, E., Millican, K., van~den Driessche, G., Lespiau, J.-B., Damoc, B., Clark, A., de~Las~Casas, D., Guy, A., Menick, J., Ring, R., Hennigan, T., Huang, S., Maggiore, L., Jones, C., Cassirer, A., Brock, A., Paganini, M., Irving, G., Vinyals, O., Osindero, S., Simonyan, K., Rae, J.~W., Elsen, E., and Sifre, L.
\newblock Improving language models by retrieving from trillions of tokens, 2022.

\bibitem[Bubeck et~al.(2023)Bubeck, Chandrasekaran, Eldan, Gehrke, Horvitz, Kamar, Lee, Lee, Li, Lundberg, Nori, Palangi, Ribeiro, and Zhang]{bubeck2023sparks}
Bubeck, S., Chandrasekaran, V., Eldan, R., Gehrke, J., Horvitz, E., Kamar, E., Lee, P., Lee, Y.~T., Li, Y., Lundberg, S., Nori, H., Palangi, H., Ribeiro, M.~T., and Zhang, Y.
\newblock Sparks of artificial general intelligence: Early experiments with gpt-4, 2023.

\bibitem[Chen et~al.(2021)Chen, Tworek, Jun, Yuan, de~Oliveira~Pinto, Kaplan, Edwards, Burda, Joseph, Brockman, Ray, Puri, Krueger, Petrov, Khlaaf, Sastry, Mishkin, Chan, Gray, Ryder, Pavlov, Power, Kaiser, Bavarian, Winter, Tillet, Such, Cummings, Plappert, Chantzis, Barnes, Herbert-Voss, Guss, Nichol, Paino, Tezak, Tang, Babuschkin, Balaji, Jain, Saunders, Hesse, Carr, Leike, Achiam, Misra, Morikawa, Radford, Knight, Brundage, Murati, Mayer, Welinder, McGrew, Amodei, McCandlish, Sutskever, and Zaremba]{humaneval}
Chen, M., Tworek, J., Jun, H., Yuan, Q., de~Oliveira~Pinto, H.~P., Kaplan, J., Edwards, H., Burda, Y., Joseph, N., Brockman, G., Ray, A., Puri, R., Krueger, G., Petrov, M., Khlaaf, H., Sastry, G., Mishkin, P., Chan, B., Gray, S., Ryder, N., Pavlov, M., Power, A., Kaiser, L., Bavarian, M., Winter, C., Tillet, P., Such, F.~P., Cummings, D., Plappert, M., Chantzis, F., Barnes, E., Herbert-Voss, A., Guss, W.~H., Nichol, A., Paino, A., Tezak, N., Tang, J., Babuschkin, I., Balaji, S., Jain, S., Saunders, W., Hesse, C., Carr, A.~N., Leike, J., Achiam, J., Misra, V., Morikawa, E., Radford, A., Knight, M., Brundage, M., Murati, M., Mayer, K., Welinder, P., McGrew, B., Amodei, D., McCandlish, S., Sutskever, I., and Zaremba, W.
\newblock Evaluating large language models trained on code, 2021.

\bibitem[Chiang et~al.(2023)Chiang, Li, Lin, Sheng, Wu, Zhang, Zheng, Zhuang, Zhuang, Gonzalez, Stoica, and Xing]{vicuna2023}
Chiang, W.-L., Li, Z., Lin, Z., Sheng, Y., Wu, Z., Zhang, H., Zheng, L., Zhuang, S., Zhuang, Y., Gonzalez, J.~E., Stoica, I., and Xing, E.~P.
\newblock Vicuna: An open-source chatbot impressing gpt-4 with 90\%* chatgpt quality, March 2023.
\newblock URL \url{https://lmsys.org/blog/2023-03-30-vicuna/}.

\bibitem[Chowdhery et~al.(2022)Chowdhery, Narang, Devlin, Bosma, Mishra, Roberts, Barham, Chung, Sutton, Gehrmann, Schuh, Shi, Tsvyashchenko, Maynez, Rao, Barnes, Tay, Shazeer, Prabhakaran, Reif, Du, Hutchinson, Pope, Bradbury, Austin, Isard, Gur-Ari, Yin, Duke, Levskaya, Ghemawat, Dev, Michalewski, Garcia, Misra, Robinson, Fedus, Zhou, Ippolito, Luan, Lim, Zoph, Spiridonov, Sepassi, Dohan, Agrawal, Omernick, Dai, Pillai, Pellat, Lewkowycz, Moreira, Child, Polozov, Lee, Zhou, Wang, Saeta, Diaz, Firat, Catasta, Wei, Meier-Hellstern, Eck, Dean, Petrov, and Fiedel]{chowdhery2022palm}
Chowdhery, A., Narang, S., Devlin, J., Bosma, M., Mishra, G., Roberts, A., Barham, P., Chung, H.~W., Sutton, C., Gehrmann, S., Schuh, P., Shi, K., Tsvyashchenko, S., Maynez, J., Rao, A., Barnes, P., Tay, Y., Shazeer, N., Prabhakaran, V., Reif, E., Du, N., Hutchinson, B., Pope, R., Bradbury, J., Austin, J., Isard, M., Gur-Ari, G., Yin, P., Duke, T., Levskaya, A., Ghemawat, S., Dev, S., Michalewski, H., Garcia, X., Misra, V., Robinson, K., Fedus, L., Zhou, D., Ippolito, D., Luan, D., Lim, H., Zoph, B., Spiridonov, A., Sepassi, R., Dohan, D., Agrawal, S., Omernick, M., Dai, A.~M., Pillai, T.~S., Pellat, M., Lewkowycz, A., Moreira, E., Child, R., Polozov, O., Lee, K., Zhou, Z., Wang, X., Saeta, B., Diaz, M., Firat, O., Catasta, M., Wei, J., Meier-Hellstern, K., Eck, D., Dean, J., Petrov, S., and Fiedel, N.
\newblock Palm: Scaling language modeling with pathways, 2022.

\bibitem[Clark et~al.(2018)Clark, Cowhey, Etzioni, Khot, Sabharwal, Schoenick, and Tafjord]{clark2018think}
Clark, P., Cowhey, I., Etzioni, O., Khot, T., Sabharwal, A., Schoenick, C., and Tafjord, O.
\newblock Think you have solved question answering? try arc, the ai2 reasoning challenge, 2018.

\bibitem[Dao et~al.(2022)Dao, Fu, Ermon, Rudra, and R{\'e}]{dao2022flashattention}
Dao, T., Fu, D.~Y., Ermon, S., Rudra, A., and R{\'e}, C.
\newblock Flash{A}ttention: Fast and memory-efficient exact attention with {IO}-awareness.
\newblock In \emph{Advances in Neural Information Processing Systems}, 2022.

\bibitem[Dong et~al.(2023)Dong, Wang, Sreedhar, Wu, and Kuchaiev]{dong2023steerlm}
Dong, Y., Wang, Z., Sreedhar, M.~N., Wu, X., and Kuchaiev, O.
\newblock Steerlm: Attribute conditioned sft as an (user-steerable) alternative to rlhf, 2023.

\bibitem[Gao et~al.(2020)Gao, Biderman, Black, Golding, Hoppe, Foster, Phang, He, Thite, Nabeshima, Presser, and Leahy]{gao2020pile}
Gao, L., Biderman, S., Black, S., Golding, L., Hoppe, T., Foster, C., Phang, J., He, H., Thite, A., Nabeshima, N., Presser, S., and Leahy, C.
\newblock The pile: An 800gb dataset of diverse text for language modeling, 2020.

\bibitem[Gao et~al.(2022)Gao, Ma, Lin, and Callan]{gao2022tevatron}
Gao, L., Ma, X., Lin, J., and Callan, J.
\newblock Tevatron: An efficient and flexible toolkit for dense retrieval, 2022.

\bibitem[Gao et~al.(2024)Gao, Xiong, Gao, Jia, Pan, Bi, Dai, Sun, Guo, Wang, and Wang]{gao2024retrievalaugmented}
Gao, Y., Xiong, Y., Gao, X., Jia, K., Pan, J., Bi, Y., Dai, Y., Sun, J., Guo, Q., Wang, M., and Wang, H.
\newblock Retrieval-augmented generation for large language models: A survey, 2024.

\bibitem[Gururangan et~al.(2020)Gururangan, Marasović, Swayamdipta, Lo, Beltagy, Downey, and Smith]{gururangan2020dont}
Gururangan, S., Marasović, A., Swayamdipta, S., Lo, K., Beltagy, I., Downey, D., and Smith, N.~A.
\newblock Don't stop pretraining: Adapt language models to domains and tasks, 2020.

\bibitem[He et~al.(2023)He, Wu, Zhang, Yao, Zheng, Zheng, and Yu]{chateda}
He, Z., Wu, H., Zhang, X., Yao, X., Zheng, S., Zheng, H., and Yu, B.
\newblock Chateda: A large language model powered autonomous agent for eda, 2023.

\bibitem[Hendrycks et~al.(2021)Hendrycks, Burns, Basart, Zou, Mazeika, Song, and Steinhardt]{hendrycks2021measuring}
Hendrycks, D., Burns, C., Basart, S., Zou, A., Mazeika, M., Song, D., and Steinhardt, J.
\newblock Measuring massive multitask language understanding, 2021.

\bibitem[Hu et~al.(2021)Hu, Shen, Wallis, Allen{-}Zhu, Li, Wang, and Chen]{LoRA}
Hu, E.~J., Shen, Y., Wallis, P., Allen{-}Zhu, Z., Li, Y., Wang, S., and Chen, W.
\newblock Lora: Low-rank adaptation of large language models.
\newblock \emph{CoRR}, abs/2106.09685, 2021.
\newblock URL \url{https://arxiv.org/abs/2106.09685}.

\bibitem[Izacard et~al.(2022{\natexlab{a}})Izacard, Caron, Hosseini, Riedel, Bojanowski, Joulin, and Grave]{contriever}
Izacard, G., Caron, M., Hosseini, L., Riedel, S., Bojanowski, P., Joulin, A., and Grave, E.
\newblock Unsupervised dense information retrieval with contrastive learning, 2022{\natexlab{a}}.

\bibitem[Izacard et~al.(2022{\natexlab{b}})Izacard, Lewis, Lomeli, Hosseini, Petroni, Schick, Dwivedi-Yu, Joulin, Riedel, and Grave]{ATLAS}
Izacard, G., Lewis, P., Lomeli, M., Hosseini, L., Petroni, F., Schick, T., Dwivedi-Yu, J., Joulin, A., Riedel, S., and Grave, E.
\newblock Atlas: Few-shot learning with retrieval augmented language models, 2022{\natexlab{b}}.

\bibitem[Ji et~al.(2023)Ji, Lee, Frieske, Yu, Su, Xu, Ishii, Bang, Madotto, and Fung]{hallucination}
Ji, Z., Lee, N., Frieske, R., Yu, T., Su, D., Xu, Y., Ishii, E., Bang, Y.~J., Madotto, A., and Fung, P.
\newblock Survey of hallucination in natural language generation.
\newblock \emph{ACM Comput. Surv.}, 55\penalty0 (12), mar 2023.
\newblock ISSN 0360-0300.
\newblock \doi{10.1145/3571730}.
\newblock URL \url{https://doi.org/10.1145/3571730}.

\bibitem[Karpukhin et~al.(2020)Karpukhin, Oğuz, Min, Lewis, Wu, Edunov, Chen, and tau Yih]{DPR}
Karpukhin, V., Oğuz, B., Min, S., Lewis, P., Wu, L., Edunov, S., Chen, D., and tau Yih, W.
\newblock Dense passage retrieval for open-domain question answering, 2020.

\bibitem[Khailany et~al.(2020)Khailany, Ren, Dai, Godil, Keller, Kirby, Klinefelter, Venkatesan, Zhang, Catanzaro, and Dally]{IEEEMicro2020}
Khailany, B., Ren, H., Dai, S., Godil, S., Keller, B., Kirby, R., Klinefelter, A., Venkatesan, R., Zhang, Y., Catanzaro, B., and Dally, W.~J.
\newblock Accelerating chip design with machine learning.
\newblock \emph{IEEE Micro}, 40\penalty0 (6):\penalty0 23--32, 2020.
\newblock \doi{10.1109/MM.2020.3026231}.

\bibitem[Kocetkov et~al.(2022)Kocetkov, Li, Allal, Li, Mou, Ferrandis, Jernite, Mitchell, Hughes, Wolf, Bahdanau, von Werra, and de~Vries]{kocetkov2022stack}
Kocetkov, D., Li, R., Allal, L.~B., Li, J., Mou, C., Ferrandis, C.~M., Jernite, Y., Mitchell, M., Hughes, S., Wolf, T., Bahdanau, D., von Werra, L., and de~Vries, H.
\newblock The stack: 3 tb of permissively licensed source code, 2022.

\bibitem[Koto et~al.(2021)Koto, Lau, and Baldwin]{koto-etal-2021-indobertweet}
Koto, F., Lau, J.~H., and Baldwin, T.
\newblock {I}ndo{BERT}weet: A pretrained language model for {I}ndonesian {T}witter with effective domain-specific vocabulary initialization.
\newblock In \emph{Proceedings of the 2021 Conference on Empirical Methods in Natural Language Processing}, pp.\  10660--10668, November 2021.

\bibitem[Kuchaiev et~al.(2019)Kuchaiev, Li, Nguyen, Hrinchuk, Leary, Ginsburg, Kriman, Beliaev, Lavrukhin, Cook, Castonguay, Popova, Huang, and Cohen]{kuchaiev2019nemo}
Kuchaiev, O., Li, J., Nguyen, H., Hrinchuk, O., Leary, R., Ginsburg, B., Kriman, S., Beliaev, S., Lavrukhin, V., Cook, J., Castonguay, P., Popova, M., Huang, J., and Cohen, J.~M.
\newblock Nemo: a toolkit for building ai applications using neural modules, 2019.

\bibitem[Köpf et~al.(2023)Köpf, Kilcher, von Rütte, Anagnostidis, Tam, Stevens, Barhoum, Duc, Stanley, Nagyfi, ES, Suri, Glushkov, Dantuluri, Maguire, Schuhmann, Nguyen, and Mattick]{köpf2023openassistant}
Köpf, A., Kilcher, Y., von Rütte, D., Anagnostidis, S., Tam, Z.-R., Stevens, K., Barhoum, A., Duc, N.~M., Stanley, O., Nagyfi, R., ES, S., Suri, S., Glushkov, D., Dantuluri, A., Maguire, A., Schuhmann, C., Nguyen, H., and Mattick, A.
\newblock Openassistant conversations -- democratizing large language model alignment, 2023.

\bibitem[Lai et~al.(2017)Lai, Xie, Liu, Yang, and Hovy]{lai2017race}
Lai, G., Xie, Q., Liu, H., Yang, Y., and Hovy, E.
\newblock Race: Large-scale reading comprehension dataset from examinations, 2017.

\bibitem[Lewis et~al.(2021{\natexlab{a}})Lewis, Perez, Piktus, Petroni, Karpukhin, Goyal, Küttler, Lewis, tau Yih, Rocktäschel, Riedel, and Kiela]{RAG}
Lewis, P., Perez, E., Piktus, A., Petroni, F., Karpukhin, V., Goyal, N., Küttler, H., Lewis, M., tau Yih, W., Rocktäschel, T., Riedel, S., and Kiela, D.
\newblock Retrieval-augmented generation for knowledge-intensive nlp tasks, 2021{\natexlab{a}}.

\bibitem[Lewis et~al.(2021{\natexlab{b}})Lewis, Perez, Piktus, Petroni, Karpukhin, Goyal, Küttler, Lewis, tau Yih, Rocktäschel, Riedel, and Kiela]{lewis2021retrievalaugmented}
Lewis, P., Perez, E., Piktus, A., Petroni, F., Karpukhin, V., Goyal, N., Küttler, H., Lewis, M., tau Yih, W., Rocktäschel, T., Riedel, S., and Kiela, D.
\newblock Retrieval-augmented generation for knowledge-intensive nlp tasks, 2021{\natexlab{b}}.

\bibitem[Lewkowycz et~al.(2022)Lewkowycz, Andreassen, Dohan, Dyer, Michalewski, Ramasesh, Slone, Anil, Schlag, Gutman-Solo, Wu, Neyshabur, Gur-Ari, and Misra]{lewkowycz2022solving}
Lewkowycz, A., Andreassen, A., Dohan, D., Dyer, E., Michalewski, H., Ramasesh, V., Slone, A., Anil, C., Schlag, I., Gutman-Solo, T., Wu, Y., Neyshabur, B., Gur-Ari, G., and Misra, V.
\newblock Solving quantitative reasoning problems with language models, 2022.

\bibitem[Liu et~al.(2023)Liu, Pinckney, Khailany, and Ren]{liu2023verilogeval}
Liu, M., Pinckney, N., Khailany, B., and Ren, H.
\newblock {VerilogEval:} evaluating large language models for verilog code generation.
\newblock In \emph{2023 IEEE/ACM International Conference on Computer-Aided Design (ICCAD)}, 2023.

\bibitem[Nijkamp et~al.(2023)Nijkamp, Pang, Hayashi, Tu, Wang, Zhou, Savarese, and Xiong]{nijkamp2022codegen}
Nijkamp, E., Pang, B., Hayashi, H., Tu, L., Wang, H., Zhou, Y., Savarese, S., and Xiong, C.
\newblock Codegen: An open large language model for code with multi-turn program synthesis.
\newblock \emph{ICLR}, 2023.

\bibitem[OpenAI et~al.(2023)OpenAI, :, Achiam, Adler, Agarwal, Ahmad, Akkaya, Aleman, Almeida, Altenschmidt, Altman, Anadkat, Avila, Babuschkin, Balaji, Balcom, Baltescu, Bao, Bavarian, Belgum, Bello, Berdine, Bernadett-Shapiro, Berner, Bogdonoff, Boiko, Boyd, Brakman, Brockman, Brooks, Brundage, Button, Cai, Campbell, Cann, Carey, Carlson, Carmichael, Chan, Chang, Chantzis, Chen, Chen, Chen, Chen, Chen, Chess, Cho, Chu, Chung, Cummings, Currier, Dai, Decareaux, Degry, Deutsch, Deville, Dhar, Dohan, Dowling, Dunning, Ecoffet, Eleti, Eloundou, Farhi, Fedus, Felix, Fishman, Forte, Fulford, Gao, Georges, Gibson, Goel, Gogineni, Goh, Gontijo-Lopes, Gordon, Grafstein, Gray, Greene, Gross, Gu, Guo, Hallacy, Han, Harris, He, Heaton, Heidecke, Hesse, Hickey, Hickey, Hoeschele, Houghton, Hsu, Hu, Hu, Huizinga, Jain, Jain, Jang, Jiang, Jiang, Jin, Jin, Jomoto, Jonn, Jun, Kaftan, Łukasz Kaiser, Kamali, Kanitscheider, Keskar, Khan, Kilpatrick, Kim, Kim, Kim, Kirchner, Kiros, Knight, Kokotajlo, Łukasz Kondraciuk,
  Kondrich, Konstantinidis, Kosic, Krueger, Kuo, Lampe, Lan, Lee, Leike, Leung, Levy, Li, Lim, Lin, Lin, Litwin, Lopez, Lowe, Lue, Makanju, Malfacini, Manning, Markov, Markovski, Martin, Mayer, Mayne, McGrew, McKinney, McLeavey, McMillan, McNeil, Medina, Mehta, Menick, Metz, Mishchenko, Mishkin, Monaco, Morikawa, Mossing, Mu, Murati, Murk, Mély, Nair, Nakano, Nayak, Neelakantan, Ngo, Noh, Ouyang, O'Keefe, Pachocki, Paino, Palermo, Pantuliano, Parascandolo, Parish, Parparita, Passos, Pavlov, Peng, Perelman, de~Avila Belbute~Peres, Petrov, de~Oliveira~Pinto, Michael, Pokorny, Pokrass, Pong, Powell, Power, Power, Proehl, Puri, Radford, Rae, Ramesh, Raymond, Real, Rimbach, Ross, Rotsted, Roussez, Ryder, Saltarelli, Sanders, Santurkar, Sastry, Schmidt, Schnurr, Schulman, Selsam, Sheppard, Sherbakov, Shieh, Shoker, Shyam, Sidor, Sigler, Simens, Sitkin, Slama, Sohl, Sokolowsky, Song, Staudacher, Such, Summers, Sutskever, Tang, Tezak, Thompson, Tillet, Tootoonchian, Tseng, Tuggle, Turley, Tworek, Uribe, Vallone,
  Vijayvergiya, Voss, Wainwright, Wang, Wang, Wang, Ward, Wei, Weinmann, Welihinda, Welinder, Weng, Weng, Wiethoff, Willner, Winter, Wolrich, Wong, Workman, Wu, Wu, Wu, Xiao, Xu, Yoo, Yu, Yuan, Zaremba, Zellers, Zhang, Zhang, Zhao, Zheng, Zhuang, Zhuk, and Zoph]{openai2023gpt4}
OpenAI, :, Achiam, J., Adler, S., Agarwal, S., Ahmad, L., Akkaya, I., Aleman, F.~L., Almeida, D., Altenschmidt, J., Altman, S., Anadkat, S., Avila, R., Babuschkin, I., Balaji, S., Balcom, V., Baltescu, P., Bao, H., Bavarian, M., Belgum, J., Bello, I., Berdine, J., Bernadett-Shapiro, G., Berner, C., Bogdonoff, L., Boiko, O., Boyd, M., Brakman, A.-L., Brockman, G., Brooks, T., Brundage, M., Button, K., Cai, T., Campbell, R., Cann, A., Carey, B., Carlson, C., Carmichael, R., Chan, B., Chang, C., Chantzis, F., Chen, D., Chen, S., Chen, R., Chen, J., Chen, M., Chess, B., Cho, C., Chu, C., Chung, H.~W., Cummings, D., Currier, J., Dai, Y., Decareaux, C., Degry, T., Deutsch, N., Deville, D., Dhar, A., Dohan, D., Dowling, S., Dunning, S., Ecoffet, A., Eleti, A., Eloundou, T., Farhi, D., Fedus, L., Felix, N., Fishman, S.~P., Forte, J., Fulford, I., Gao, L., Georges, E., Gibson, C., Goel, V., Gogineni, T., Goh, G., Gontijo-Lopes, R., Gordon, J., Grafstein, M., Gray, S., Greene, R., Gross, J., Gu, S.~S., Guo, Y.,
  Hallacy, C., Han, J., Harris, J., He, Y., Heaton, M., Heidecke, J., Hesse, C., Hickey, A., Hickey, W., Hoeschele, P., Houghton, B., Hsu, K., Hu, S., Hu, X., Huizinga, J., Jain, S., Jain, S., Jang, J., Jiang, A., Jiang, R., Jin, H., Jin, D., Jomoto, S., Jonn, B., Jun, H., Kaftan, T., Łukasz Kaiser, Kamali, A., Kanitscheider, I., Keskar, N.~S., Khan, T., Kilpatrick, L., Kim, J.~W., Kim, C., Kim, Y., Kirchner, H., Kiros, J., Knight, M., Kokotajlo, D., Łukasz Kondraciuk, Kondrich, A., Konstantinidis, A., Kosic, K., Krueger, G., Kuo, V., Lampe, M., Lan, I., Lee, T., Leike, J., Leung, J., Levy, D., Li, C.~M., Lim, R., Lin, M., Lin, S., Litwin, M., Lopez, T., Lowe, R., Lue, P., Makanju, A., Malfacini, K., Manning, S., Markov, T., Markovski, Y., Martin, B., Mayer, K., Mayne, A., McGrew, B., McKinney, S.~M., McLeavey, C., McMillan, P., McNeil, J., Medina, D., Mehta, A., Menick, J., Metz, L., Mishchenko, A., Mishkin, P., Monaco, V., Morikawa, E., Mossing, D., Mu, T., Murati, M., Murk, O., Mély, D., Nair, A.,
  Nakano, R., Nayak, R., Neelakantan, A., Ngo, R., Noh, H., Ouyang, L., O'Keefe, C., Pachocki, J., Paino, A., Palermo, J., Pantuliano, A., Parascandolo, G., Parish, J., Parparita, E., Passos, A., Pavlov, M., Peng, A., Perelman, A., de~Avila Belbute~Peres, F., Petrov, M., de~Oliveira~Pinto, H.~P., Michael, Pokorny, Pokrass, M., Pong, V., Powell, T., Power, A., Power, B., Proehl, E., Puri, R., Radford, A., Rae, J., Ramesh, A., Raymond, C., Real, F., Rimbach, K., Ross, C., Rotsted, B., Roussez, H., Ryder, N., Saltarelli, M., Sanders, T., Santurkar, S., Sastry, G., Schmidt, H., Schnurr, D., Schulman, J., Selsam, D., Sheppard, K., Sherbakov, T., Shieh, J., Shoker, S., Shyam, P., Sidor, S., Sigler, E., Simens, M., Sitkin, J., Slama, K., Sohl, I., Sokolowsky, B., Song, Y., Staudacher, N., Such, F.~P., Summers, N., Sutskever, I., Tang, J., Tezak, N., Thompson, M., Tillet, P., Tootoonchian, A., Tseng, E., Tuggle, P., Turley, N., Tworek, J., Uribe, J. F.~C., Vallone, A., Vijayvergiya, A., Voss, C., Wainwright, C.,
  Wang, J.~J., Wang, A., Wang, B., Ward, J., Wei, J., Weinmann, C., Welihinda, A., Welinder, P., Weng, J., Weng, L., Wiethoff, M., Willner, D., Winter, C., Wolrich, S., Wong, H., Workman, L., Wu, S., Wu, J., Wu, M., Xiao, K., Xu, T., Yoo, S., Yu, K., Yuan, Q., Zaremba, W., Zellers, R., Zhang, C., Zhang, M., Zhao, S., Zheng, T., Zhuang, J., Zhuk, W., and Zoph, B.
\newblock Gpt-4 technical report, 2023.

\bibitem[Ouyang et~al.(2022)Ouyang, Wu, Jiang, Almeida, Wainwright, Mishkin, Zhang, Agarwal, Slama, Ray, Schulman, Hilton, Kelton, Miller, Simens, Askell, Welinder, Christiano, Leike, and Lowe]{ouyang2022training}
Ouyang, L., Wu, J., Jiang, X., Almeida, D., Wainwright, C.~L., Mishkin, P., Zhang, C., Agarwal, S., Slama, K., Ray, A., Schulman, J., Hilton, J., Kelton, F., Miller, L., Simens, M., Askell, A., Welinder, P., Christiano, P., Leike, J., and Lowe, R.
\newblock Training language models to follow instructions with human feedback, 2022.

\bibitem[Pearce et~al.(2020)Pearce, Tan, and Karri]{dave}
Pearce, H., Tan, B., and Karri, R.
\newblock Dave: Deriving automatically verilog from english.
\newblock In \emph{Proceedings of the 2020 ACM/IEEE Workshop on Machine Learning for CAD}, MLCAD '20, pp.\  27–32, New York, NY, USA, 2020. Association for Computing Machinery.
\newblock ISBN 9781450375191.
\newblock \doi{10.1145/3380446.3430634}.
\newblock URL \url{https://doi.org/10.1145/3380446.3430634}.

\bibitem[Rafailov et~al.(2023)Rafailov, Sharma, Mitchell, Ermon, Manning, and Finn]{rafailov2023direct}
Rafailov, R., Sharma, A., Mitchell, E., Ermon, S., Manning, C.~D., and Finn, C.
\newblock Direct preference optimization: Your language model is secretly a reward model, 2023.

\bibitem[Ram et~al.(2023)Ram, Levine, Dalmedigos, Muhlgay, Shashua, Leyton-Brown, and Shoham]{incontext-RAG}
Ram, O., Levine, Y., Dalmedigos, I., Muhlgay, D., Shashua, A., Leyton-Brown, K., and Shoham, Y.
\newblock In-context retrieval-augmented language models, 2023.

\bibitem[Reimers \& Gurevych(2019)Reimers and Gurevych]{reimers-2019-sentence-bert}
Reimers, N. and Gurevych, I.
\newblock Sentence-bert: Sentence embeddings using siamese bert-networks.
\newblock In \emph{Proceedings of the 2019 Conference on Empirical Methods in Natural Language Processing}. Association for Computational Linguistics, 11 2019.
\newblock URL \url{http://arxiv.org/abs/1908.10084}.

\bibitem[Ren \& Fojtik(2021)Ren and Fojtik]{NVCell}
Ren, H. and Fojtik, M.
\newblock Invited- nvcell: Standard cell layout in advanced technology nodes with reinforcement learning.
\newblock In \emph{2021 58th {ACM}/{IEEE} Design Automation Conference ({DAC})}, 2021.

\bibitem[Richardson(2007)]{BeautifulSoup}
Richardson, L.
\newblock Beautiful soup documentation.
\newblock \emph{April}, 2007.

\bibitem[Robertson \& Zaragoza(2009)Robertson and Zaragoza]{BM25}
Robertson, S. and Zaragoza, H.
\newblock The probabilistic relevance framework: Bm25 and beyond.
\newblock \emph{Found. Trends Inf. Retr.}, 3\penalty0 (4):\penalty0 333–389, apr 2009.
\newblock ISSN 1554-0669.
\newblock \doi{10.1561/1500000019}.
\newblock URL \url{https://doi.org/10.1561/1500000019}.

\bibitem[Roy et~al.(2021)Roy, Raiman, Kant, Elkin, Kirby, Siu, Oberman, Godil, and Catanzaro]{prefixRL}
Roy, R., Raiman, J., Kant, N., Elkin, I., Kirby, R., Siu, M., Oberman, S., Godil, S., and Catanzaro, B.
\newblock {PrefixRL}: Optimization of parallel prefix circuits using deep reinforcement learning.
\newblock In \emph{2021 58th {ACM}/{IEEE} Design Automation Conference ({DAC})}, 2021.

\bibitem[Rozière et~al.(2023)Rozière, Gehring, Gloeckle, Sootla, Gat, Tan, Adi, Liu, Remez, Rapin, Kozhevnikov, Evtimov, Bitton, Bhatt, Ferrer, Grattafiori, Xiong, Défossez, Copet, Azhar, Touvron, Martin, Usunier, Scialom, and Synnaeve]{rozière2023code}
Rozière, B., Gehring, J., Gloeckle, F., Sootla, S., Gat, I., Tan, X.~E., Adi, Y., Liu, J., Remez, T., Rapin, J., Kozhevnikov, A., Evtimov, I., Bitton, J., Bhatt, M., Ferrer, C.~C., Grattafiori, A., Xiong, W., Défossez, A., Copet, J., Azhar, F., Touvron, H., Martin, L., Usunier, N., Scialom, T., and Synnaeve, G.
\newblock Code llama: Open foundation models for code, 2023.

\bibitem[Sakaguchi et~al.(2019)Sakaguchi, Bras, Bhagavatula, and Choi]{sakaguchi2019winogrande}
Sakaguchi, K., Bras, R.~L., Bhagavatula, C., and Choi, Y.
\newblock Winogrande: An adversarial winograd schema challenge at scale.
\newblock \emph{arXiv preprint arXiv:1907.10641}, 2019.

\bibitem[Sanh et~al.(2022)Sanh, Webson, Raffel, Bach, Sutawika, Alyafeai, Chaffin, Stiegler, Scao, Raja, Dey, Bari, Xu, Thakker, Sharma, Szczechla, Kim, Chhablani, Nayak, Datta, Chang, Jiang, Wang, Manica, Shen, Yong, Pandey, Bawden, Wang, Neeraj, Rozen, Sharma, Santilli, Fevry, Fries, Teehan, Bers, Biderman, Gao, Wolf, and Rush]{sanh2022multitask}
Sanh, V., Webson, A., Raffel, C., Bach, S.~H., Sutawika, L., Alyafeai, Z., Chaffin, A., Stiegler, A., Scao, T.~L., Raja, A., Dey, M., Bari, M.~S., Xu, C., Thakker, U., Sharma, S.~S., Szczechla, E., Kim, T., Chhablani, G., Nayak, N., Datta, D., Chang, J., Jiang, M. T.-J., Wang, H., Manica, M., Shen, S., Yong, Z.~X., Pandey, H., Bawden, R., Wang, T., Neeraj, T., Rozen, J., Sharma, A., Santilli, A., Fevry, T., Fries, J.~A., Teehan, R., Bers, T., Biderman, S., Gao, L., Wolf, T., and Rush, A.~M.
\newblock Multitask prompted training enables zero-shot task generalization, 2022.

\bibitem[Shi et~al.(2023)Shi, Min, Yasunaga, Seo, James, Lewis, Zettlemoyer, and tau Yih]{REPLUG}
Shi, W., Min, S., Yasunaga, M., Seo, M., James, R., Lewis, M., Zettlemoyer, L., and tau Yih, W.
\newblock Replug: Retrieval-augmented black-box language models, 2023.

\bibitem[Shoeybi et~al.(2019)Shoeybi, Patwary, Puri, LeGresley, Casper, and Catanzaro]{shoeybi2019megatron}
Shoeybi, M., Patwary, M., Puri, R., LeGresley, P., Casper, J., and Catanzaro, B.
\newblock Megatron-lm: Training multi-billion parameter language models using model parallelism.
\newblock \emph{arXiv preprint arXiv:1909.08053}, 2019.

\bibitem[Taylor et~al.(2022)Taylor, Kardas, Cucurull, Scialom, Hartshorn, Saravia, Poulton, Kerkez, and Stojnic]{taylor2022galactica}
Taylor, R., Kardas, M., Cucurull, G., Scialom, T., Hartshorn, A., Saravia, E., Poulton, A., Kerkez, V., and Stojnic, R.
\newblock Galactica: A large language model for science, 2022.

\bibitem[Thakur et~al.(2023)Thakur, Ahmad, Fan, Pearce, Tan, Karri, Dolan-Gavitt, and Garg]{benchmark}
Thakur, S., Ahmad, B., Fan, Z., Pearce, H., Tan, B., Karri, R., Dolan-Gavitt, B., and Garg, S.
\newblock Benchmarking large language models for automated verilog rtl code generation.
\newblock In \emph{2023 Design, Automation \& Test in Europe Conference \& Exhibition (DATE)}, pp.\  1--6, 2023.
\newblock \doi{10.23919/DATE56975.2023.10137086}.

\bibitem[Touvron et~al.(2023)Touvron, Martin, Stone, Albert, Almahairi, Babaei, Bashlykov, Batra, Bhargava, Bhosale, Bikel, Blecher, Ferrer, Chen, Cucurull, Esiobu, Fernandes, Fu, Fu, Fuller, Gao, Goswami, Goyal, Hartshorn, Hosseini, Hou, Inan, Kardas, Kerkez, Khabsa, Kloumann, Korenev, Koura, Lachaux, Lavril, Lee, Liskovich, Lu, Mao, Martinet, Mihaylov, Mishra, Molybog, Nie, Poulton, Reizenstein, Rungta, Saladi, Schelten, Silva, Smith, Subramanian, Tan, Tang, Taylor, Williams, Kuan, Xu, Yan, Zarov, Zhang, Fan, Kambadur, Narang, Rodriguez, Stojnic, Edunov, and Scialom]{touvron2023llama}
Touvron, H., Martin, L., Stone, K., Albert, P., Almahairi, A., Babaei, Y., Bashlykov, N., Batra, S., Bhargava, P., Bhosale, S., Bikel, D., Blecher, L., Ferrer, C.~C., Chen, M., Cucurull, G., Esiobu, D., Fernandes, J., Fu, J., Fu, W., Fuller, B., Gao, C., Goswami, V., Goyal, N., Hartshorn, A., Hosseini, S., Hou, R., Inan, H., Kardas, M., Kerkez, V., Khabsa, M., Kloumann, I., Korenev, A., Koura, P.~S., Lachaux, M.-A., Lavril, T., Lee, J., Liskovich, D., Lu, Y., Mao, Y., Martinet, X., Mihaylov, T., Mishra, P., Molybog, I., Nie, Y., Poulton, A., Reizenstein, J., Rungta, R., Saladi, K., Schelten, A., Silva, R., Smith, E.~M., Subramanian, R., Tan, X.~E., Tang, B., Taylor, R., Williams, A., Kuan, J.~X., Xu, P., Yan, Z., Zarov, I., Zhang, Y., Fan, A., Kambadur, M., Narang, S., Rodriguez, A., Stojnic, R., Edunov, S., and Scialom, T.
\newblock Llama 2: Open foundation and fine-tuned chat models, 2023.

\bibitem[Venigalla et~al.(2022)Venigalla, Frankle, and Carbin]{biomed}
Venigalla, A., Frankle, J., and Carbin, M.
\newblock {BioMedLM}: a domain-specific large language model for biomedical text, 2022.
\newblock URL \url{https://www.mosaicml.com/blog/introducing-pubmed-gpt}.

\bibitem[Wang et~al.(2022)Wang, Yang, Huang, Jiao, Yang, Jiang, Majumder, and Wei]{e5}
Wang, L., Yang, N., Huang, X., Jiao, B., Yang, L., Jiang, D., Majumder, R., and Wei, F.
\newblock Text embeddings by weakly-supervised contrastive pre-training.
\newblock \emph{arXiv preprint arXiv:2212.03533}, 2022.

\bibitem[Wang et~al.(2023)Wang, Dong, Zeng, Adams, Sreedhar, Egert, Delalleau, Scowcroft, Kant, Swope, and Kuchaiev]{wang2023helpsteer}
Wang, Z., Dong, Y., Zeng, J., Adams, V., Sreedhar, M.~N., Egert, D., Delalleau, O., Scowcroft, J.~P., Kant, N., Swope, A., and Kuchaiev, O.
\newblock Helpsteer: Multi-attribute helpfulness dataset for steerlm, 2023.

\bibitem[Wei et~al.(2022)Wei, Bosma, Zhao, Guu, Yu, Lester, Du, Dai, and Le]{wei2022finetuned}
Wei, J., Bosma, M., Zhao, V.~Y., Guu, K., Yu, A.~W., Lester, B., Du, N., Dai, A.~M., and Le, Q.~V.
\newblock Finetuned language models are zero-shot learners, 2022.

\bibitem[Wu et~al.(2023)Wu, Irsoy, Lu, Dabravolski, Dredze, Gehrmann, Kambadur, Rosenberg, and Mann]{wu2023bloomberggpt}
Wu, S., Irsoy, O., Lu, S., Dabravolski, V., Dredze, M., Gehrmann, S., Kambadur, P., Rosenberg, D., and Mann, G.
\newblock Bloomberggpt: A large language model for finance, 2023.

\bibitem[Zellers et~al.(2019)Zellers, Holtzman, Bisk, Farhadi, and Choi]{zellers2019hellaswag}
Zellers, R., Holtzman, A., Bisk, Y., Farhadi, A., and Choi, Y.
\newblock Hellaswag: Can a machine really finish your sentence?
\newblock In \emph{Proceedings of the 57th Annual Meeting of the Association for Computational Linguistics}, 2019.

\bibitem[Zhou et~al.(2023)Zhou, Alon, Xu, Wang, Jiang, and Neubig]{docprompting}
Zhou, S., Alon, U., Xu, F.~F., Wang, Z., Jiang, Z., and Neubig, G.
\newblock Docprompting: Generating code by retrieving the docs, 2023.

\end{thebibliography}
\bibliographystyle{icml2024}

\appendix
\clearpage
\section{Appendix}
\subsection{Contributions}

\textbf{Mingjie Liu} conducted DAPT and model alignment.

\textbf{Teodor-Dumitru Ene, Robert Kirby} developed inference and application evaluation infrastructure. 

\textbf{Chris Cheng} developed RAG framework.

\textbf{Nathaniel Pinckney} collected and prepared data sets for training.

\textbf{Rongjian Liang} developed custom tokenizers.

\textbf{Walker Turner, Charley Lind, George Kokai} developed a general circuit design knowledge benchmark. 

\textbf {Siddhanth Dhodhi, Ismet Bayraktaroglu, Himyanshu Anand, Eric Hill} designed engineering assistant chatbot, provided domain instruction datasets, evaluation benchmarks, and conducted evaluation.

\textbf{Parikshit Deshpande, Zhengjiang Shao, Kaizhe Xu, Jiashang Hu, Laura Dang, Xiaowei Li, Hao Liu, Ambar Sarkar} developed engineering assistant chatbot application.

\textbf{Sreedhar Pratty, Kishor Kunal, Ghasem Pasandi, Varun Tej, Sumit Jain, Sujeet Omar, Pratik P Suthar, Hanfei Sun} developed EDA scripts generation application, provided domain instruction datasets and evaluation benchmarks. 

\textbf{Bonita Bhaskaran, Arjun Chaudhuri, Sanmitra Banerjee, Ghasem Pasandi} developed bug summarization and analysis application, provided domain instruction datasets and evaluation benchmarks.

\textbf{Brucek Khailany, Stuart Oberman, Sharon Clay, Sameer Halepete, Jonathan Raiman, Bryan Catanzaro, Jonah Alben, Bill Dally} advised from AI research and hardware engineering perspectives.

\textbf{Haoxing Ren} designed and led the research.

\subsection{Data Collection Process}
\label{app:data}
Collection was implemented with a set of shell and Python scripts, designed to identify relevant design data and documentation, convert them to plain text if applicable, filter them using basic quality metrics, compute a checksum for precise file deduplication, and compress them for storage. The collection flow did not use off-the-shelf LLM-specific scraping and collection scripts, as we aimed to minimize space requirements through in-situ data collection of internal data sources (both networked file systems and internal web applications). For file system-based collection, data was kept in-place while being filtered for quality, instead of storing additional sets of raw data locally.

The design and verification data collection encompassed a variety of source files, including Verilog and VHDL (RTL and netlists), C++, Spice, Tcl, various scripting languages, and build-related configuration files. Data from internal web services were gathered through both REST API calls and conventional crawling, with HTML formatting being removed using the open-source BeautifulSoup\cite{BeautifulSoup} Python library in both instances to minimize inadvertent removal of coding examples, at the cost of introducing more boiler plate navigation bars and other HTML page elements. Our data collection flow supported conventional documentation formats, including .docx, .pptx, and .pdf, using readily available Python conversion libraries and open-source tools.

As most internal data is believe to be of high quality, minimal filtering was applied: line count filtering was used to ensure that exceedingly large or small files were excluded, and files were sorted into broad categories of manually written versus tool-generated. 

\label{sec:dataset}

\subsection{Training Data}
\label{sec:appendix:data}

During Domain-Adaptive Pre-Training (DAPT), we assemble a dataset from a combination of proprietary chip design specific data sources and publicly available datasets.

\textbf{Chip Design Datasets: }
Our internal dataset consists of a diverse range of text sources pertinent to chip design, spanning design, verification, infrastructure, and internal documentation. Table~\ref{tab:train_tokens} provides a breakdown of the data collected after filtering, and the corresponding number of tokens using the LLaMA2 tokenizer. We construct the dataset by gathering all relevant internal data, then filtering by file type, based on filename extensions and distinguishing between machine-generated and human-written content.
Although we evaluated on three specific use cases, we did not specifically limit the dataset to sources known to be relevant to these use cases since we believed that incorporating additional domain knowledge would improve performance. 
After collection, cleaning, and filtering, the internal data training corpus has 23.1 billion tokens. 
Further details of the data collection process are covered in Appendix~\ref{app:data}.

\begin{table*}[ht]
    \small
    \centering
    \begin{tabular}{|l|c|c|c|c|}
        \hline
        \multirow{2}{*}{Data Source Type}   & Data  &   Data  & Training & Training\\
                                            & Percentage (\%) &   Tokens (B) & Percentage (\%) &   Tokens (B) \\
        \hline
        \hline
        Bug Summary & 9.5\% & 2.4 & 10.0\% & 2.4 \\
        Design Source & 47.0\% & 11.9 & 24.5\% & 5.9  \\
        Documentation & 17.8\% & 4.5 & 34.0\% & 8.2  \\
        Verification & 9.1\% & 2.3 & 10.4\% & 2.5  \\
        Other & 7.9\% & 2.0 & 12.0\% & 2.9  \\
        \hline
        Wikipedia & 5.9\% & 1.5 & 6.2\% & 1.5 \\
        Github & 2.8\% & 0.7 & 3.0\% & 0.7 \\
        \hline
        \hline
        Total & 100.0\% & 25.3 & 100.0\% & 24.1 \\
        \hline
    \end{tabular}
    \caption{Breakdown of Data by Source. Token count measured with original LLaMA2 tokenizer.}
    \label{tab:train_tokens}
\end{table*}

\textbf{Public Datasets:} 
We augment the chip design specific data with a sample of publicly available data from various sources, a common practice in the development of foundational large language models. Our approach was to reuse public training data from other language models, with the stipulation that it must be publicly accessible and compatible with open sourcing. These datasets exhibit a high degree of correlation with the pretraining data used in LLaMA2~\cite{touvron2023llama}, with the intention of preserving general knowledge and natural language capabilities during DAPT.
The public datasets used by ChipNeMo can be categorized into two groups, natural language and code. For the natural language component, we draw from Wikipedia data~\cite{gao2020pile}, as it is widely regarded for its high data quality. For code, we leverage GitHub data~\cite{kocetkov2022stack}, focusing on programming languages also present in our internal data chip design dataset such as C++, Python, and Verilog. To ensure that the overall dataset is representative of pre-training distributions, we perform a sub-sampling operation that results in approximately 9.2\% of the total training tokens being sampled from these public datasets, with a balanced representation of natural language and code.

\textbf{Data Blend:} 
A significant proportion of the domain data we gathered is comprised of unannotated code from diverse origins. In an effort to enhance the model's comprehension of domain-specific knowledge, we conducted downsampling of code data while concurrently upsampling natural language data, specifically design documentation, over a span of 2 to 4 training epochs. We also increased the representation of data that we deemed more pertinent to downstream applications, such as human-written EDA tool scripts. Furthermore, we incorporated publicly available domain data for 1 epoch. Details of the token distribution for training are shown in Table~\ref{tab:train_tokens}. 

\subsection{Alignment Data}
\label{sec:appendix:data:sft}
During Supervised Fine-Tuning (SFT), we employ a general chat SFT instruction dataset that is accessible for commercial use. The dataset is comprised largely of publicly available instruction following datasets including OASST~\cite{köpf2023openassistant}, FLAN~\cite{wei2022finetuned}, P3~\cite{sanh2022multitask} and a small amount of a broad domain proprietary dataset comprising various topics such as brainstorming, open-ended question answering, rewriting, summarization etc. It's important to note that the SFT instruction data we discuss here is focused on general natural language tasks and does not contain any information or tasks related to the downstream use cases in chip design. In total, this dataset comprises 128,000 training samples.

For SteerLM~\cite{dong2023steerlm} we closely follow the implementations in~\cite{wang2023helpsteer}. The attribute training data only contains public available data from HelpSteer~\cite{wang2023helpsteer} and OASST~\cite{köpf2023openassistant}. For the models attribute-conditioned finetuning,we only used the OASST data comprised of 56,000 training samples.

Additionally, we meticulously assembled a domain-specific instruction dataset for aligning the model to downstream use cases. These examples have been meticulously crafted by subject matter experts and are formatted as single-turn questions and answers. Table~\ref{tab:sft} depicts the quantity of our domain-specific instruction dataset. It's worth noting that the total number of training samples in the domain-specific instruction dataset is quite small when compared to the extensive amount of generative chat instruction data.

\begin{table}[ht]
    \small
    \centering
    \begin{tabular}{|l|c|}
        \hline
        Domain Source & Number of Samples \\
        \hline
        Design Knowledge & 302 \\
        EDA Script Generation & 480 \\
        Bug summarization and analysis & 648 \\
        \hline
        Total & 1430 \\
        \hline
    \end{tabular}
    \caption{Breakdown of Domain Alignment Data.}
    \label{tab:sft}
\end{table}

\subsection{Domain Evaluation Benchmarks} 
\label{sec:appendix:evaluation}

In order to quickly and quantitatively assess the accuracy of various models, we established evaluation criteria structured as multiple-choice question-and-answer formats for each use case, designed to closely align with established benchmarks, such as MMLU~\cite{hendrycks2021measuring}. In the process of formulating these multiple-choice questions, collaboration with domain experts was pivotal. The goal was to ensure that each question included at least one complex answer choice, thereby posing a challenge to individuals with limited domain expertise. Careful attention was also given to prevent any inadvertent contamination of the questions with data from our domain-specific alignment data.  In addition to the per-use-case benchmarks, an additional benchmark was created for general circuit design knowledge, covering both analog and digital design topics. The number of multiple-choice questions for evaluation benchmark are shown in Table~\ref{tab:autoeval}.

\begin{table}[ht]
    \scriptsize
    \centering
    \begin{tabular}{|l|c|}
        \hline
        Domain Source & Number of Questions \\
        \hline
        Design Knowledge (Design)  & 94 \\
        EDA Script Generation (Scripting) & 74 \\
        Bug Summarization and Analysis (Bugs) & 70 \\
        Open Domain Circuit Design (Circuits) & 227 \\
        \hline
    \end{tabular}
    \caption{Domain-specific Evaluation Benchmark.}
    \label{tab:autoeval}
\end{table}

When we report results on the above benchmarks, we take average results obtained from five distinct runs to mitigate the effects of variance and noise in the testing process. Each iteration employs a set of 5-shot examples, with variations introduced across each individual runs.

In addition to these domain-specific evaluation benchmarks, we also include commonly-used publicly available LLM academic benchmarks. Furthermore, we measure the model's code generation capabilities, by evaluating HumanEval~\cite{humaneval} for Python and VerilogEval~\cite{liu2023verilogeval} for Verilog.

\subsection{Domain Adaptive Pretraining (DAPT)} 
\label{sec:training}

In this section we present detailed results on our domain adaptive pretrained models. We also detail our ablation experiments on domain adaptive pretraining.

\textbf{DAPT Hyperparameters: } Details presented in Table~\ref{tab:hyperparams}.

\begin{table}[ht]
    \small
    \centering
    \begin{tabular}{|c|c|}
        \hline
        Hyperparameters & Value \\
        \hline
        Context Window & 4096 \\
        Global Batch Size & 256 (128) \\
        Optimizer & distributed\_fused\_adam \\
        Weight Decay & 0.01 \\
        Betas & 0.9, 0.95 (0.9, 0.98) \\
        Learning Rate & $5 \cdot 10^{-6}$ \\
        Scheduler & None \\
        \hline
    \end{tabular}
    \caption{DAPT and SteerLM/SFT hyperparameters, SteerLM/SFT values shown in parenthesis (if differs from DAPT).}
    \label{tab:hyperparams}
\end{table}

\textbf{Auto Eval Results:}
We present detailed results on auto evaluation benchmarks in Table~\ref{tab:auto_evals_detail} and Table~\ref{tab:coding_results}. For simplicity, in the remainders of the section we present aggregated benchmark results for ablation studies:
\begin{itemize}
    \item \textbf{Chip:} We report average results on in-domain Design, Scripting, Bugs, and Circuits benchmarks from Table~\ref{tab:autoeval} (5-shot).
    \item \textbf{MMLU:} We report the overall results on MMLU (5-shot)~\cite{hendrycks2021measuring} a popular aggregated benchmark on a wide variety of subjects.
    \item \textbf{Reasoning:} We report average results on popular public benchmarks on common sense reasoning (0-shot), including Winogrande~\cite{sakaguchi2019winogrande}, hellaswag~\cite{zellers2019hellaswag}, ARC-easy~\cite{clark2018think}, and RACE-High~\cite{lai2017race}.
    \item \textbf{Code:} We report average pass-rate of coding benchmarks with greedy decoding, including HumanEval~\cite{humaneval}, VerilogEval-Machine~\cite{liu2023verilogeval}, and VerilogEval-Human~\cite{liu2023verilogeval}.
\end{itemize}

\begin{table*}[ht]
    \small
    \centering
    \begin{tabular}{|p{2.2cm}|C{1.0cm}|C{1.0cm}|C{1.0cm}|C{1.0cm}|C{1.0cm}|C{1.4cm}|C{1.2cm}|C{1.0cm}|C{1.2cm}|}
        \hline
        {Model} & {Design} &  {Scripting} & {Bugs} & {Circuits} & MMLU & Winogrande & hellaswag & ARC-e & RACE-H \\
        \hline
        LLaMA2-7B & 41.1 & 42.0 & 42.2 & 47.9 & 45.7 & 68.9 & 75.6 & 73.5 & 46.2\\
        ChipNeMo-7B & 57.5 & 49.3 & 42.8 & 49.5  & 44.6 & 67.4 & 76.3 & 73.7 & 46.2 \\
        \hline
        LLaMA2-13B & 43.6 & 49.6 & 39.7 & 55.5 & 55.4 & 72.1 & 79.3 & 76.3 & 46.7\\
        ChipNeMo-13B & 67.9 & 56.3 & 50.1 & 56.8 & 53.4 & 71.1 & 80.3 & 76.7 & 46.1\\
        \hline
        LLaMA2-70B & 52.3 & 64.9 & 56.9 & 67.0 & 68.6 & 77.6 & 83.6 & 79.6 & 48.9 \\
        ChipNeMo-70B & 76.6 & 73.9 & 65.8 & 71.7 & 69.4 & 78.0 & 85.1 & 80.9 & 48.9 \\
        \hline \hline
        GPT-3.5 & 51.7 & 66.7 & 52.0 & 66.5 & $70.0^*$ & $81.6^*$ & $85.5^*$ & $85.2^*$ & - \\
        GPT-4  & 58.4 & 77.4 & 63.4 & 79.0 & $86.4^*$ & $87.5^*$ & $95.3^*$ & $96.3^*$ & - \\
        \hline
    \end{tabular}
    \caption{Auto Evaluation Results. We report academic benchmark results for LLaMA2 using proprietary evaluation methods. ChipNeMo models trained with domain-adapted tokenizer. * results from~\cite{openai2023gpt4}.}
    \label{tab:auto_evals_detail}
\end{table*}

\begin{table}[ht]
    \small
    \centering
    \begin{tabular}{|p{2.2cm}|C{1.4cm}|C{1.4cm}|C{1.4cm}|}
        \hline
        \multirow{2}{*}{Model} & \multirow{2}{*}{HumanEval} & VerilogEval-Human & VerilogEval-Machine \\
        \hline
        LLaMA2-7B & 14.0 & 3.8 & 24.5 \\
        ChipNeMo-7B & 12.2 & 8.3 & 28.7 \\
        \hline
        LLaMA2-13B & 17.1 & 9.0 & 30.8 \\
        ChipNeMo-13B & 17.7 & 22.4 & 43.4 \\
        \hline
        LLaMA2-70B & 28.0 & 30.8 & 51.0 \\
        ChipNeMo-70B & 30.5 & 27.6 & 53.8 \\
        \hline \hline
        GPT-3.5 & $48.1^*$ & $26.7^{\dag}$ & $46.7^{\dag}$\\
        GPT-4 & $67.0^*$ & $43.5^{\dag}$ & $60.0^{\dag}$\\
        \hline
    \end{tabular}
    \caption{Coding Evaluation Results. Showing pass-rate with greedy decoding. We report results for LLaMA2 using proprietary evaluation methods. ChipNeMo models trained with domain-adapted tokenizer. *, \dag~results from~\cite{openai2023gpt4, liu2023verilogeval}.}
    \label{tab:coding_results}
\end{table}

\begin{figure}[h]
    \vspace{-0.1in}
    \centering
    \includegraphics[width=.42\textwidth]{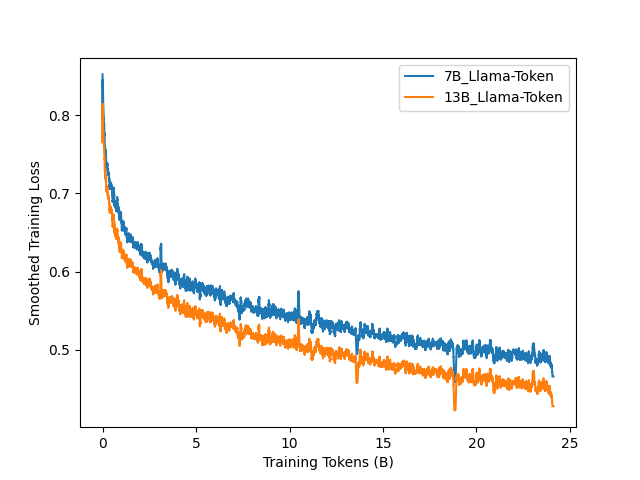}
    \caption{Smoothed Training Loss with Original LLaMA2 Tokenizer.}
    \label{fig:tokenizer_loss}
    \vspace{-0.1in}
\end{figure}

\textbf{Domain-Adaptive Tokenization:}
We experimented with DAPT using the original LLaMA2 tokenizer and the domain-adapted tokenizer as described in Section~\ref{sec:tokenizer}. Figure~\ref{fig:tokenizer_loss} depicts smoothed training loss for ChipNeMo with the original unmodified tokenizer. When compared with Figure~\ref{fig:loss_llama2}, we observe that the domain-adapted tokenizer has larger training loss upon initialization, due to added tokens never being observed during foundation model pretraining. Similar training loss is achieved for DAPT with 1 epoch. 

Table~\ref{tab:tokenizer_ablation} presents aggregated auto evaluation benchmark results. We note that careful tokenizer adaptations and weight initialization only slightly impacts model performance on general academic benchmarks. DAPT significantly improved domain benchmarks with any tokenizer, including Verilog coding (no major difference in HumanEval). We conclude that domain-adapting the tokenizer comes with the benefit of improved tokenization and training efficiency with no degradation on the models general language and domain capabilities.

\begin{table}[ht]
    \scriptsize		
    \centering
    \begin{tabular}{|C{0.6cm}|C{1.0cm}|C{0.6cm}|C{0.6cm}|C{0.65cm}|C{0.6cm}|C{0.6cm}|}
        \hline
        Model & Tokenizer & DAPT & Chip & MMLU & Reason & Code \\
        \hline
        7B & Ori. & No &  43.4 & 45.7 & 66.1 & 14.1\\
        7B & Dpt. & No & 42.7 & 44.6 & 65.9 & 13.9\\
        7B & Ori. & Yes & 51.2 & 44.8 & 65.7 & 17.6\\
        7B & Dpt. & Yes & 49.8 & 44.6 & 65.8 & 16.4\\
        \hline
        13B & Ori. & No & 47.1 & 55.4 & 68.6 & 18.9\\
        13B & Dpt. & No & 46.0 & 55.1 & 68.6 & 18.4\\
        13B & Ori. & Yes & 57.7 & 54.0 & 68.4 & 27.2\\
        13B & Dpt. & Yes & 57.8 & 53.4 & 68.5 & 27.8\\
        \hline
    \end{tabular}
    \caption{Evaluation Results on ChipNeMo models with Different Tokenizers. \texttt{Dpt.} indicate domain-adapted tokenizer and \texttt{Ori.} indicate using LLaMA2 original tokenizer. Using augmented tokenizer without DAPT corresponds to the model initialization as in Section~\ref{sec:tokenizer}.}
    \label{tab:tokenizer_ablation}
\end{table}

\textbf{Public Datasets Mix-in:}
As introduced in Section~\ref{sec:appendix:data} we included public data in DAPT, sampled from commonly-used public datasets for foundation model pre-training. We primarily hoped that mixing in public data such as Wikipedia in DAPT could help ``correct'' disturbances brought by domain-adapted tokenizer and improve general natural language capabilities of models. We conducted another round of DAPT with domain-adapted tokenizer using only the domain data, training for the same number of steps equating to roughly 1.1 epoch of the data. We found that public data mix-in slightly improves results. We present detailed results in Table~\ref{tab:ablation_public_data}.

\begin{table}[ht]
    \small
    \centering
    \begin{tabular}{|C{0.8cm}|C{0.8cm}|C{0.9cm}|C{0.8cm}|C{0.8cm}|C{0.8cm}|}
        \hline
         Public & Chip & MMLU & Reason & Code  \\
        \hline
         No &  56.9 & 53.0 & 67.5 & 24.1\\
         Yes &  57.8 & 53.4 & 68.5 & 27.8\\

        \hline
    \end{tabular}
    \caption{Ablation on Public Dataset Mix-in with ChipNeMo-13B. Public data mix-in slightly improves results.}
    \label{tab:ablation_public_data}
\end{table}

\textbf{Learning Rate:}
We experimented with employing a larger learning rate, inspired by the approach used in \mbox{CodeLLaMA}~\cite{rozière2023code}. We use similar training hyperparameters as in~Table~\ref{tab:hyperparams_large}. We use a cosine schedule with 200 warm-up steps, and set the final learning rate to be 1/30th of the peak learning rate of $3 \cdot 10^{-4}$. We use the same batch size and number of training steps as DAPT. 

\begin{table}[ht]
    \small
    \centering
    \begin{tabular}{|c|c|}
        \hline
        Hyperparameters & Value \\
        \hline
        Context Window & 4096 \\
        Global Batch Size & 256  \\
        Optimizer & distributed\_fused\_adam \\
        Weight Decay & 0.01 \\
        Betas & 0.9, 0.95 \\
        Learning Rate (lr) & $3 \cdot 10^{-4}$ \\
        Scheduler & CosineAnnealing \\
        Warmup Steps & 200 \\
        min\_lr & $1 \cdot 10^{-5}$ \\
        \hline
    \end{tabular}
    \caption{Training Hyperparameters with Larger Learning Rate. We adopt similar parameter as to~\cite{rozière2023code}.}
    \label{tab:hyperparams_large}
\end{table}

\begin{figure}[t]
    \vspace{-0.1in}
    \centering
    \includegraphics[width=.42\textwidth]{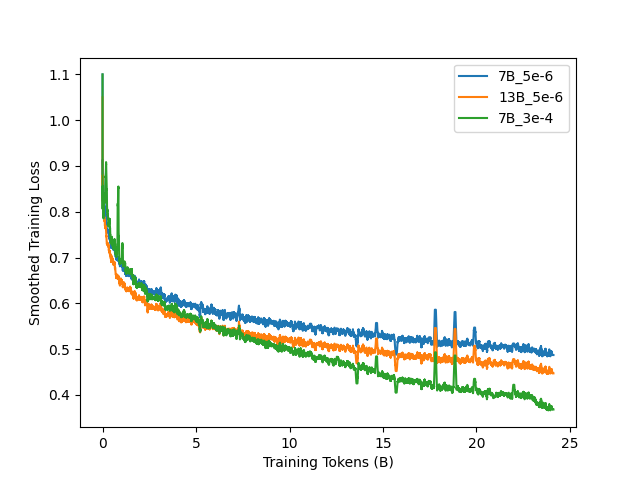}
    \caption{Smoothed Training Loss with Larger Learning Rate. We include loss curves of suggested hyperparameters for comparison.}
    \label{fig:loss_large_lr}
\end{figure}

Figure~\ref{fig:loss_large_lr} shows the training loss for ChipNeMo-7B with augmented tokenizers including public dataset mix-in. We observed large spikes in training loss at the initial training steps with the final training loss for 7B models to even be better than 13B original DAPT hyperparameters. However, we note substantial degradation across natural language benchmarks as shown in Table~\ref{tab:autoevals_large}, including in-domain chip design. Coding capabilities improved as consistent with the findings of~\cite{rozière2023code}.

\begin{table}[ht]
    \small
    \centering
    \begin{tabular}{|C{1.2cm}|C{0.8cm}|C{0.9cm}|C{0.8cm}|C{0.8cm}|C{0.8cm}|}
        \hline
         Learning Rate & Chip & MMLU & Reason & Code  \\
        \hline
         $5 \cdot 10^{-6}$ &  49.8 & 44.6 & 65.8 & 16.4 \\
         $3 \cdot 10^{-4}$ &  25.5 & 26.6 & 49.8 & 18.1\\
        \hline
    \end{tabular}
    \caption{Ablation on Learning Rate with ChipNeMo-7B. A larger learning rate significantly degrades performance on all language related tasks but slightly improves coding.}
    \label{tab:autoevals_large}
\end{table}

We highlight that our case differs from that in~\cite{rozière2023code}. Although we also conduct ``continued pretraining'' initializing from pretrained checkpoints, we preferably want the model to maintain high degrees of performance on general capabilities, while distilling domain dataset information and knowledge (unseen in model pretraining) into model weights. In contrast,~\cite{rozière2023code} use publicly available code data that predominantly lacks natural language elements, emphasizing their primary focus on coding-related tasks. We hypothesize that a smaller learning rate played a dual role for domain adaptation, facilitating the distillation of domain knowledge through DAPT while maintaining a balance that did not veer too far from the base model, thus preserving general natural language capabilities while significantly improving performance on in-domain tasks.

\textbf{Parameter Efficient Fine-Tuning (PEFT): }
Parameter efficient fine-tuning freezes the pre-trained model weights and injects trainable parameters in smaller adapter models for efficient fine-tuning of downstream tasks. We explore the use of PEFT in DAPT using Low-Rank Adaptation (LoRA)~\cite{LoRA}. Since our transformer layer implementation fuses KQV into a single projection, we add LoRA adapters for a single Low-Rank projection for each self attention layer in combined fashion. We experiment on LLaMA2-13B models with the original LLaMA2 tokenizer, using the same DAPT training setups in Table~\ref{tab:hyperparams}. We ran two experiments, introducing additional trainable parameters of 26.4 million (small) and 211.2 million (large) respectively. 

\begin{figure}[t]
    \vspace{-0.1in}
    \centering
    \includegraphics[width=.42\textwidth]{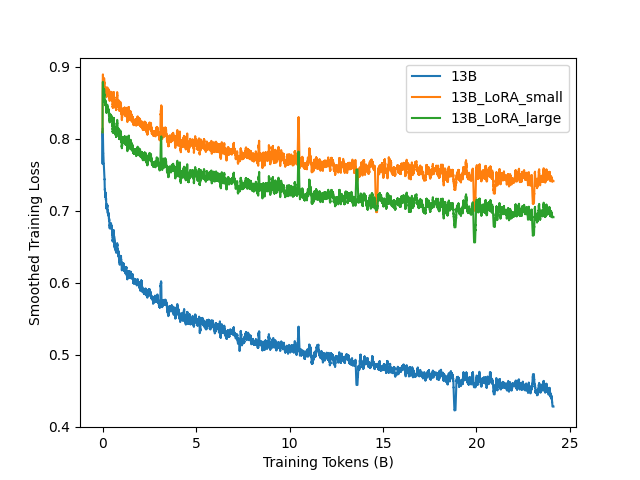}
    \caption{Smoothed Training Loss of LoRA~\cite{LoRA}. 13B corresponds to full parameter DAPT.}
    \label{fig:loss_peft}
    \vspace{-0.1in}
\end{figure}

Figure~\ref{fig:loss_peft} shows the training loss curves of LoRA models and compares with full parameter training. For both LoRA models, the loss quickly converges and stops decreasing beyond a certain point. Table~\ref{tab:autoevals_peft} reports the evaluation results on LoRA models. Both LoRA models significantly underperforms full parameter training on in-domain chip design tasks. LoRA models improve in chip design tasks compared to their non-DAPT counterparts, with the larger model exhibiting slightly better (but non significant) results.

\begin{table}[ht]
    \small
    \centering
    \begin{tabular}{|C{1.6cm}|C{0.8cm}|C{0.8cm}|C{0.8cm}|C{0.8cm}|C{0.8cm}|}
        \hline
         Parameters & Chip & MMLU & Reason & Code  \\
        \hline
         None & 47.1 & 55.4 & 68.6 & 18.9 \\
         26.4M &  49.0 & 55.0 & 68.2 & 13.0\\
         211.2M &  49.6 & 54.2 & 68.6 & 15.3\\
         13B &  57.7 & 54.0 & 68.4 & 27.2 \\
        \hline
    \end{tabular}
    \caption{Evaluation Results on LoRA Models. First column indicate number of trainable parameters. None indicates LLaMA2-13B model without DAPT. 13B indicates full parameter training.}
    \label{tab:autoevals_peft}
\end{table}


\subsection{Model Alignment}
\label{sec:appendix:training:sft}
For standard supervised-finetuning (SFT) we used the following structured template:
\begin{lstlisting}
<extra_id_0>System\n{system}
<extra_id_1>User\n{user_utterance}
<extra_id_1>Assistant\n{chipnemo_response}
...
\end{lstlisting}

For SteerLM we follow the steps in~\cite{wang2023helpsteer} and apply attribute labeling to our domain data:
\begin{enumerate}
    \item We trained a "general" attribute scoring model. We only used HelpSteer and OASST attribute labeled data without any domain data with weights initialized from LLaMA2-13B.
    \item We scored domain data (1.4k samples) with the attribute scoring model in Step 1. 
    \item We mixed OASST data (56k samples) with domain data (1.4k samples) for 2 epochs.
    \item We conduct attribute-conditioned fine-tuning on ChipNeMo models. 
\end{enumerate}

\begin{figure}[ht]
    \centering
    \includegraphics[trim={1in 0in 1in 0in},clip,width=0.45\textwidth]{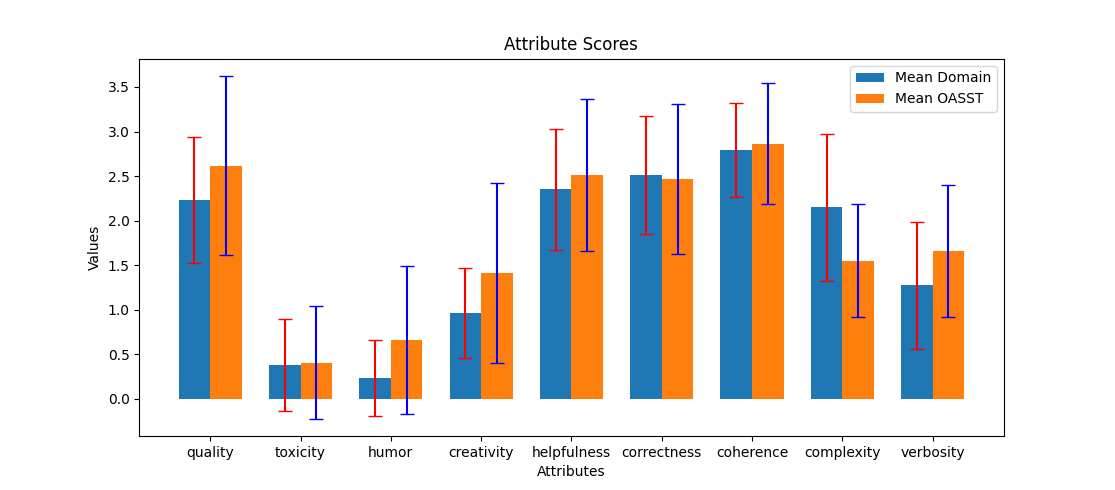}
    \caption{Attribute Scores for SteerLM.}
    \label{fig:steerlm_attributes}
\end{figure}

Figure~\ref{fig:steerlm_attributes} depicts the attribute scores (and their standard deviation) labeled by the general attribute scoring model. The attribute scoring model could generalize well to unseen domain data on attributes such as \textit{toxicity, humor, creativity, verbosity}, but had slightly lower scores for domain data on metrics such as \textit{quality, helpfulness, correctness}. This leaves room of improvement for the attribute scoring model on domain data, which could be improved by possibly initializing the attribute models from ChipNeMo (as respect to LLaMA2) and adding attribute labeled domain data. For our experiments we used the attribute labels as-is.

Additionally, we conducted an additional alignment using solely the general chat dataset, excluding any domain-specific alignment data. For clarity, we designate all our ChipNeMo models as follows:
\begin{itemize}
    \item \textbf{ChipNeMo-SFT$^{G}$:} Models fine-tuned with general chat data exclusively using standard SFT.
    \item \textbf{ChipNeMo-SFT:} Models fine-tuned with both domain and general chat data using standard SFT;
    \item \textbf{ChipNeMo-Steer$^{G}$:} Models fine-tuned with general chat data exclusively using SteerLM.
    \item \textbf{ChipNeMo-Steer:} Models fine-tuned with both domain and general chat data using SteerLM;
\end{itemize}

\subsection{Domain-Adaptive Retrieval Model}
\label{sec:retrieval_model_training}
Manually generating training samples is very effort intensive, so we elected to implement a process to generate them automatically. Since we are using contrastive learning to fine-tune our model, each sample requires a set of both positive passages and negative passages, particularly hard negatives to maximize the accuracy.

\begin{figure}[th]
    \vspace{-0.1in}
    \centering
    \includegraphics[trim={1.0in 5.5in 1.6in 0.5in},clip, width=.35\textwidth]{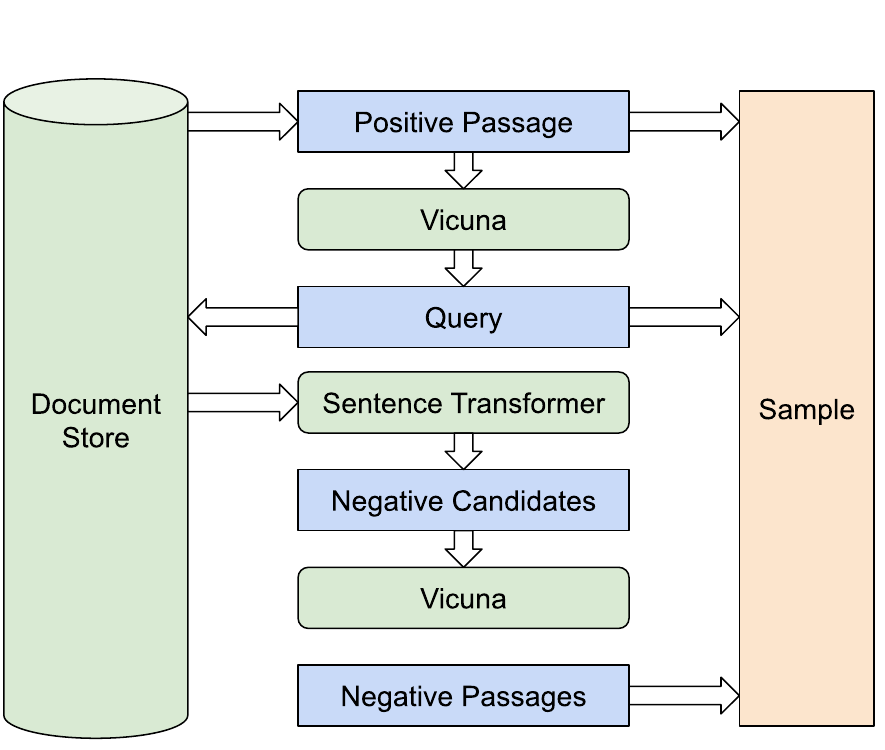}
    \caption{Sample Generation For Retrieval Model Training}
    \label{fig:retrieval_samples}
    \vspace{-0.1in}
\end{figure}

\subsubsection{Dataset Sampling Procedure}
Figure \ref{fig:retrieval_samples} describes the steps taken to generate a sample:
\begin{enumerate}
    \item Randomly select a passage from the document corpus
\item Use a language model (Vicuna) to generate a valid query from the passage
\item Use a pre-existing retrieval model (sentence transformer) to fetch the top-N passages from the document corpus for the query where each passage is a potential hard-negative
\item It is possible that some of the fetched passages are actually positive, so use the same language model to filter out the positive passages
\item If there are not enough negative passages after this filtering process, supplement with random passages from the corpus

\end{enumerate}

For our initial research we used Vicuna~\cite{vicuna2023} and Sentence Transformer~\cite{reimers-2019-sentence-bert}; however, they can easily be replaced with LLaMA2~\cite{touvron2023llama} and BM25~\cite{BM25} respectively to produce a retrieval model that is commercially viable.

\subsubsection{Hit Quality Comparison}
Not all hits are created equal. The passage in the Spec example below clearly and completely answers its query. The passage in the Build example contains the answer; however, more context is required to answer the query.


\textbf{Spec Example:} Hit passage clearly answers the query.

\begin{tcolorbox}[width=1.0\linewidth, halign=left, colframe=black, colback=white, boxsep=0.01mm, arc=1.5mm, left=2mm, right=2mm, boxrule=0.5pt]\footnotesize
\textbf{Query:}\\
What is the XXX buffer in the YYY used for? \\

\vspace{0.2in}
\textbf{Hit Passage:}\\ 
Since the memory subsystem is partitioned, YYY may receive read responses whose order is different from their requests. ZZZ requires strict in-order per-thread processing, thus YYY reorders the read responses to the request order before sending them back to ZZZ. YYY re-orders the read responses in a read return reorder buffer (XXX) to guarantee the ordering.\\

\vspace{0.2in}
\textbf{Answer:} re-orders the read responses.\\

\end{tcolorbox}

\textbf{Build Example:} Additional information is required to fully answer the query. Such as: What is a DL? How do we know Arch-Build-Hotseat-XXX is a DL?

\begin{tcolorbox}[width=1.0\linewidth, halign=left, colframe=black, colback=white, boxsep=0.01mm, arc=1.5mm, left=2mm, right=2mm, boxrule=0.5pt]\footnotesize
\textbf{Query:}\\
What is the support DL for XXX build issues?\\

\vspace{0.2in}
\textbf{Hit Passage:}\\ 
Tree Setup\\
Working in the XXX Mainline explains initial tree setup and build steps\\
Build\\
Arch-Build-Hotseat-XXX - Hotseat support for XXX build issues YYY build failures\\

\end{tcolorbox}


\subsection{Additional Evaluation Results and Discussions}
\label{sec:appendix:additional_eval}

Table~\ref{tab:full_eval} shows the evaluation data for all models on the engineering assistant chatbot application. 
It should be noted that adding domain-specific instructional data to our SFT training led to only marginal improvement on the engineering assistant task. However, including this data in our SteerLM training degraded the quality of our model's responses. We believe this is due to insufficient adaptation of our data to the SteerLM labeling system. Future work will see closer compatibility between general-purpose and domain-specific instructional data labeling, from which we expect to see improvements in the behavior of the model as a chatbot.

The evaluation results for all models on the EDA script generation task are presented in Table~\ref{tab:full_vividpt_eval}. In addition to the comparison with off-the-shelf models discussed in the main section of the paper \ref{sec:vivid}, an ablation study was conducted to assess the significance of SteerLM training against ChipNeMo-70B-SFT and ChipNeMo-70B-Steer models. The observations suggest that SteerLM training helps on the ``hard'' benchmark, thus showcasing its effectiveness in generating relevant results for real world use cases.

\begin{figure}[ht]
    \centering
    \includegraphics[trim={0in 0in 0in 0in},clip, width=.45\textwidth]{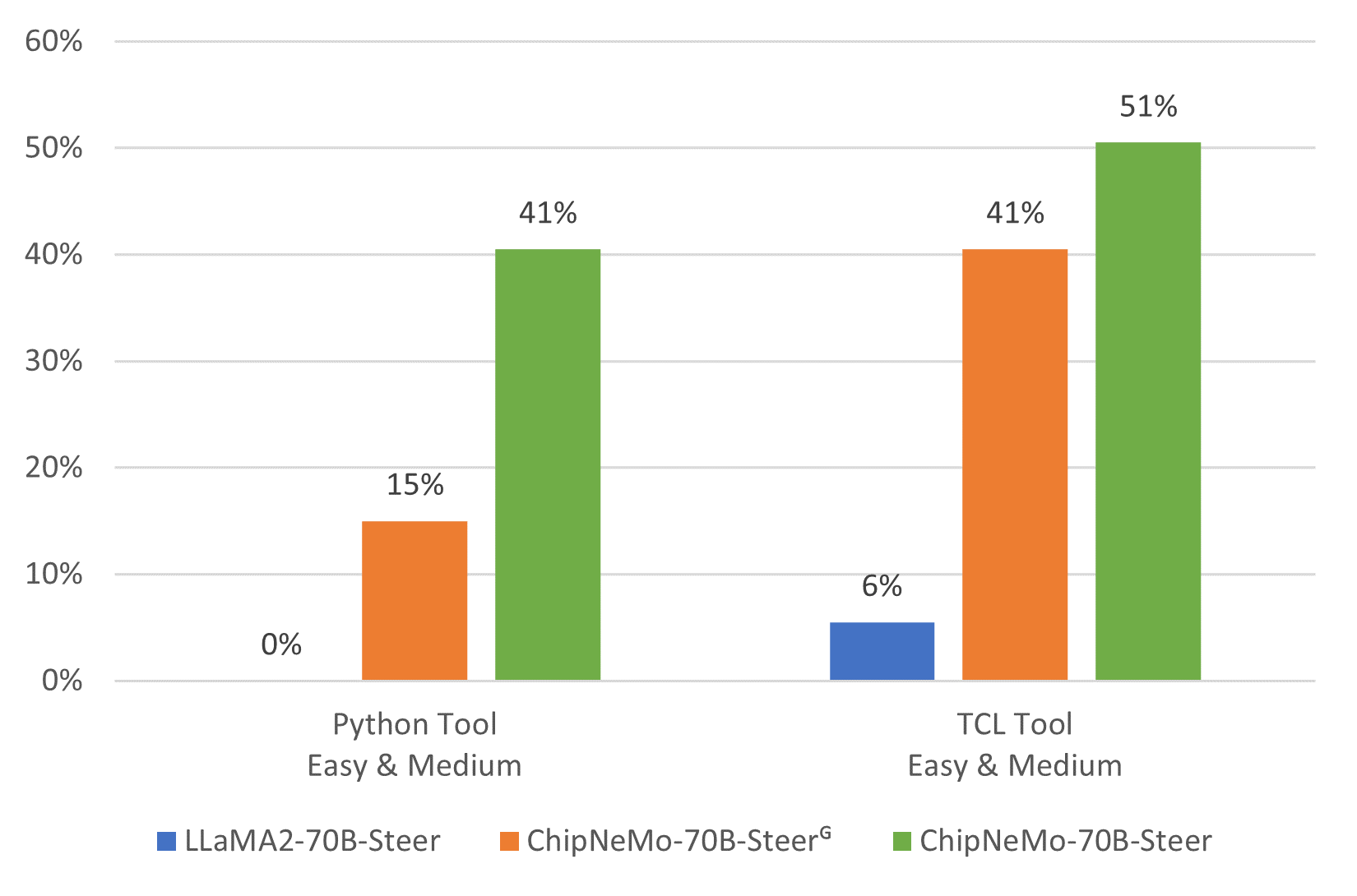}
    \caption{EDA Script Generation Domain Data Ablation, Pass@5}
    \label{fig:eda_ablation1}
\end{figure}
\begin{figure}[ht]
    \centering
    \includegraphics[trim={0in 0in 0in 0in},clip, width=.45\textwidth]{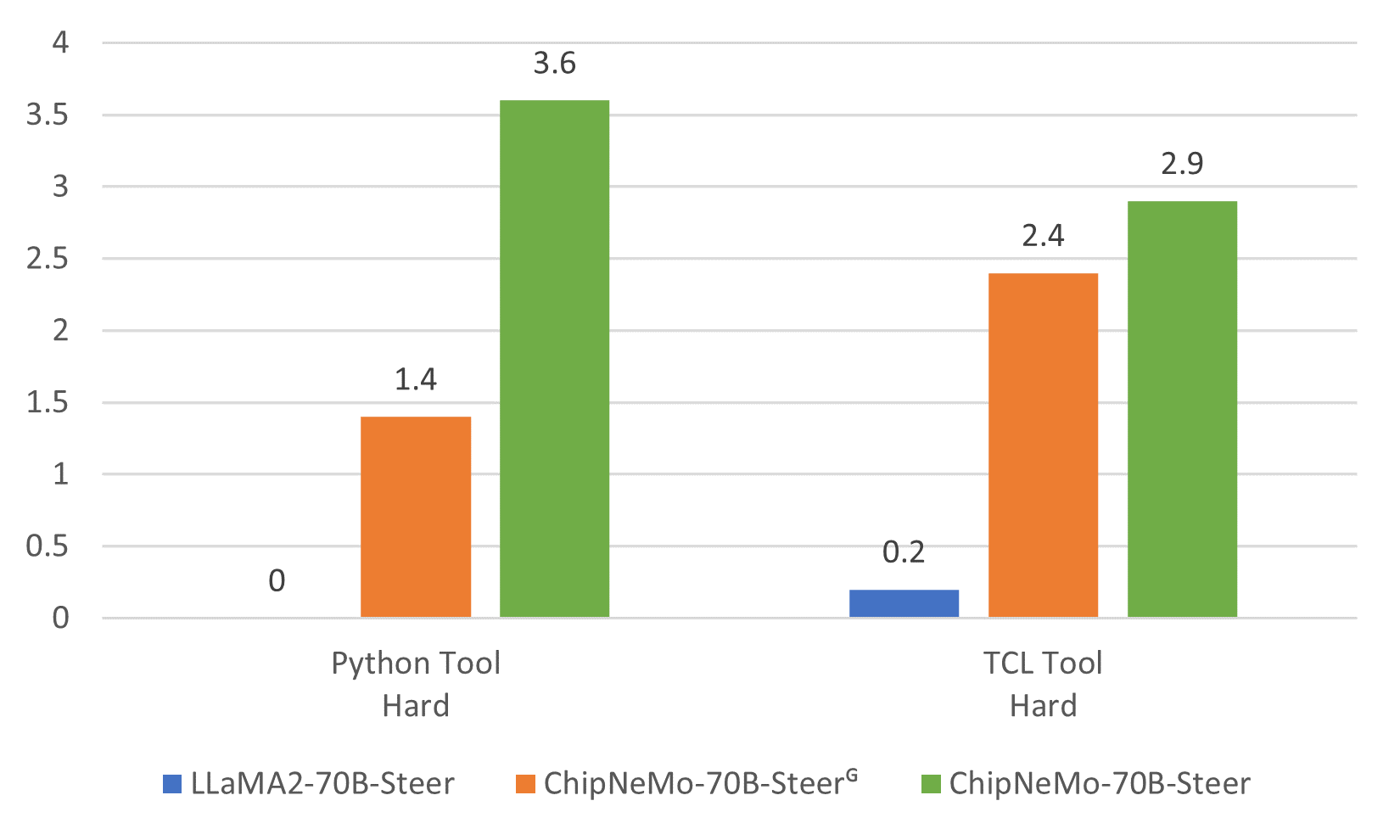}
    \caption{EDA Script Generation Domain Data Ablation, Single Generation (temperature=0), Human Evaluated 0-10}
    \label{fig:eda_ablation2}
\end{figure}

Figure \ref{fig:eda_ablation1} and Figure \ref{fig:eda_ablation2} depict the comparison among LLaMA2-70B-Steer, ChipNeMo-70B-Steer$^G$ model, and ChipNeMo-70B-Steer models. The results reveal a significant enhancement achieved solely through DAPT training, emphasizing the crucial role of domain knowledge. Moreover, a substantial improvement is observed when contrasting the chat model's performance with and without domain instructional data. This emphasizes that the model's capacity to produce accurate answers can be enhanced through improved alignment with domain instructional data. These results show the importance of both DAPT and model alignment for domain specific applications.

The impact of RAG for generating EDA scripts was also studied. The retrieved data consisted of specific APIs related to the questions along with a description of the API. This helped in improving the accuracy of ``easy'' and ``medium'' difficulty benchmark which depend heavily on the API knowledge.
On the ``hard'' benchmark, a degradation in the accuracy is noticed as compared to non-RAG model. This showcases the difficulty of coupling existing retrieval techniques with non-natural language tasks such as code generation. While a wide variety of retrieval techniques have been proven to work for natural language tasks, there are comparatively fewer publications focused on retrieval-augment code generation ~\cite{gao2024retrievalaugmented}. This emphasizes the importance of DAPT, especially in data regimes where there are insufficient quality examples and explanations to readily apply retrieval.


The evaluation results for the bug summarization and analysis task are presented in Table~\ref{tab:full_nvbugs_eval}. Minor improvements are observed with DAPT, evident when comparing the non-DAPT LLaMA2-70B-Steer$^G$ model to its DAPT counterpart, ChipNeMo-70B-Steer$^G$. When comparing SteerLM to traditional SFT training, a slight enhancement in summarization task performance is noted, while no significant difference is observed in task assignment.

Additionally, we investigated GPT-4 models in two versions. In the first approach, we utilized the conventional evaluation procedure employed for our other models, assuming limitations based on a comparable context window size. This procedure included breaking down substantial bugs into smaller segments and employing hierarchical summarization, as previously described. In the second approach, we successfully fitted each of our test bugs entirely within the 32k context size of GPT-4. The results of this ablation study suggest that our hierarchical summarization has minimal impact on response quality.



\label{app:chat}
\begin{table}[ht]
    \footnotesize
    \centering
    \begin{tabular}{|p{3.1cm}|c|c|c|c|c|c|c|c|}
        \hline
        Model & RAG & Hit & Miss & Avg. \\
        \hline
        \hline
        GPT-4 & No & - & - & 2.84 \\
        LLaMA2-70B-Chat & No & - & - & 1.81 \\
        LLaMA2-70B-Steer$^{G}$ & No & - & - & 1.99 \\
        \hline
        ChipNeMo-70B-Steer$^{G}$ & No & - & - & \bf{5.12} \\
        ChipNeMo-70B-Steer & No & - & - & 4.80 \\
        ChipNeMo-70B-SFT$^{G}$ & No & - & - & 4.68 \\
        ChipNeMo-70B-SFT & No & - & - & 4.89 \\
        \hline
        \hline
        GPT-4 & Yes & 4.77 & 4.04 & 4.52 \\
        LLaMA2-70B-Chat & Yes & 4.18 & 3.22 & 3.86 \\
        LLaMA2-70B-Steer$^{G}$ & Yes & 3.68 & 3.00 & 3.46 \\
        \hline
        ChipNeMo-70B-Steer$^{G}$ & Yes & \bf{6.02} & 4.78 & \bf{5.68} \\
        ChipNeMo-70B-Steer & Yes & 5.58 & 4.39 & 5.26 \\
        ChipNeMo-70B-SFT$^{G}$ & Yes & 5.18 & \bf{4.79} & 5.06 \\
        ChipNeMo-70B-SFT & Yes & 5.39 & 4.17 & 5.06 \\
        \hline
        
        \hline
    \end{tabular}

    \caption{Engineering Assistant Chatbot Human Evaluation. Evaluated with 7-point Likert Scale.}
    \label{tab:full_eval}
\end{table}

\begin{table*}[ht]
    \small
    \centering
    \begin{tabular}{|l|c|c|c|c|c|c|}
        \hline
        & \multicolumn{3}{c|}{Tool1 (Python)} & \multicolumn{3}{c|}{Tool2 (Tcl)} \\
        \cline{2-7}
        Model & Easy & Medium & Hard & Easy & Medium & Hard \\
        & (Automatic)  & (Automatic)  & (Human) & (Automatic) & (Automatic) & (Human)  \\
        \hline
        GPT-4 & 0\% & 0\% & 0.0 & 20\% & 52\% & 1.1  \\ 
        LLaMA2-70B-Chat & 0\% & 0\% & 0.1 & 7\% & 4\% & 0.0  \\ 
        LLaMA2-70B-Steer$^{G}$ & 0\% & 0\% & 0 & 0\% & 11\% & 0.2 \\ 
        \hline
        ChipNeMo-70B-Steer$^{G}$ & 19\% & 11\% & 1.4 & 29\% & 52\% & 2.4 \\ 
        ChipNeMo-70B-SFT & 61\% & 29\% & 3.4 & 27\% & 74\% & 1.9 \\ 
        ChipNeMo-70B-Steer & 49\% & 32\% & \textbf{3.6} & 45\% & 56\% & \textbf{2.9}  \\ 
        \hline
        ChipNeMo-70B-Steer (w/RAG) & \textbf{77\%} & \textbf{36\%} & 2.3 & \textbf{84\%} & \textbf{85\%} & 0.8 \\ 
        \hline
    \end{tabular}
    \captionsetup{justification=centering}
    \caption{EDA Script Generation Evaluation.\hspace{\linewidth}Automatic Evaluation Scored \textbf{Pass@5}.\hspace{\linewidth}Human Evaluation Scored 0-10 on a Single Generation (temperature = 0).}
    \label{tab:full_vividpt_eval}
\end{table*}

\begin{table*}[ht]
    \small
    \centering
    \begin{tabular}{|l|c|c|c|}
        \hline
        Model & Technical Summary & Managerial Summary & Task Assignment \\
        \hline
        GPT-4 & \bf{6.30} & 6.25 & \bf{6.00} \\ 
        GPT-4 (32k, No Chunks) & 6.14 & \bf{6.45} & 5.78 \\ 
        \hline
        LLaMA2-70B-Chat & 4.35 & 4.95 & 5.00 \\ 
        LLaMA2-70B-Steer$^{G}$ & 4.95 & 5.35 & 4.73 \\ 
        \hline
        ChipNeMo-70B-Steer$^{G}$ & 5.00 & 5.35 & 5.45 \\ 
        ChipNeMo-70B-Steer & 5.05 & 5.25 & 4.27 \\ 
        ChipNeMo-70B-SFT & 4.50 & 5.15 & 5.45 \\ 
        \hline
    \end{tabular}
    \caption{Bug Summarization and Analysis Evaluation. Likert scale 1-7.}
    \label{tab:full_nvbugs_eval}
\end{table*}

\subsection{Chip Design Applications}
\label{sec:application}
 We conducted a survey of potential LLM applications within our design teams and categorized them into four buckets: \textbf{code generation}, \textbf{question \& answer}, \textbf{analysis and reporting}, and \textbf{triage}. Code generation refers to LLM generating design code, testbenches, assertions, internal tools scripts, etc.; Q \& A refers to an LLM answering questions about designs, tools, infrastructures, etc.; Analysis and reporting refers to an LLM analyzing data and providing reports; triage refers to an LLM helping debug design or tool problems given logs and reports. We selected one key application from each category to study in this work, except for the \textbf{triage} category which we leave for further research. The motivation and technical details of each application are given below.

\subsubsection{Engineering Assistant Chatbot}
This application aims to help design engineers with answers to their architecture, design, verification, and build questions, which could significantly improve their overall productivity without impacting the productivity of others. It is observed that design engineers often enjoy brainstorming, designing hardware, and writing code, but can be slowed down waiting for answers on design knowledge they lack. Design productivity can also be enhanced by avoiding having engineers write code based on mistaken assumptions or debugging code that they are unfamiliar with. 
Internal studies have shown that up to 60\% of a typical chip designer's time is spent in debug or checklist related tasks across a range of topics including design specifications, testbench construction, architecture definition, and tools or infrastructure. Experts on these issues are often spread around the globe in a multinational company, such that it is not always convenient to find immediate help. Therefore, an engineering assistant chatbot based on knowledge extracted from internal design documents, code, any recorded data about designs and technical communications such as emails and corporate instant communications, etc. could help significantly improve design productivity. We implemented this application with the domain-adapted RAG method mentioned in Section~\ref{sec:rag}.

\subsubsection{EDA Script Generation}
Another common task in an industrial chip design flow is writing EDA scripts to accomplish a variety of tasks such as design implementation, introspection and transformation. 
These scripts often leverage both tool-specific and custom internal script libraries. Learning these libraries, navigating tool documentation, and writing and debugging these scripts, can take up a significant amount of engineering time.

LLMs have proven adept at small scale code generation on a wide array of tasks \cite{rozière2023code} and therefore customizing these models to accelerate engineer productivity in this domain specific task is a natural fit. 
In this work we focus on generating two different types of scripts from natural language task descriptions.
The first are scripts which leverage an internal python library for design editing and analysis. The second are Tcl scripts that use the command interface provided by a leading industrial static timing analysis tool. 

In order to build our domain-specific fine-tuning dataset for this task, production scripts for both tools were collected from design experts. We observed that our DAPT models can generate reasonable inline comments for the code. This enabled us to use these models to improve the quality of collected scripts by generating additional inline comments. Human experts later verified and corrected these comments and created an associated prompt. These prompts and code pairs make up the data used for model alignment as discussed in~\ref{sec:appendix:data:sft}.

\begin{figure}[t]
    \centering
    \includegraphics[width=0.45\textwidth]{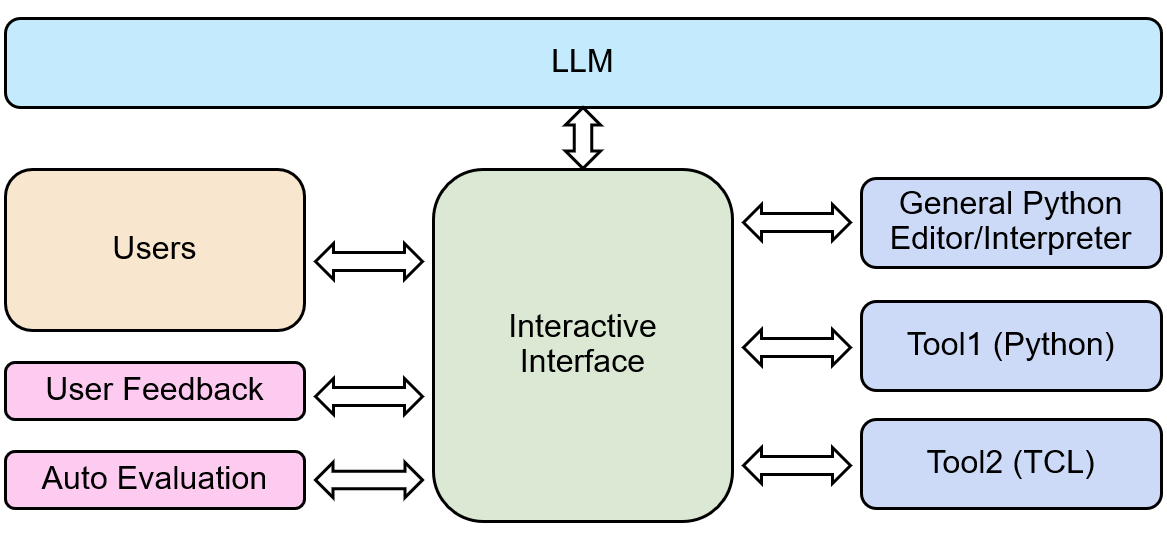}
    \caption{LLM script generator integration with EDA tools}
    \label{fig:tool_integration}
\end{figure}

To provide and collect feedback in the most meaningful way, we spent significant effort building the flow shown in Fig.~\ref{fig:tool_integration} where engineers can both query the model and run generated code through the same interface. This allows us to be confident in the \textit{correctness} of generated code as well as provide accurate feedback by allowing engineers to see how many corrections they might need to get a functioning script. We support this integration by establishing interactive connections to tool servers.

Additionally, we provide a user feedback form, allowing us to compare different models and glean valuable insights from user feedback. This valuable information can aid us in further refining our models.

\subsubsection{Bug Summarization and Analysis}
\label{subsubsec:nvbugs}
Tracking the reporting, triage, debug and resolution of various features and bugs across stages of the production flow is a time-consuming process. Engineering managers spend a lot of time reviewing internal issue tracking databases to build understanding of the state of the project and help speed their execution. Therefore, a tool that is able to look at all supporting information and quickly summarize both technical and managerial data as well as suggest next steps would boost team productivity. We focus on using LLMs to generate three different outputs - one focused on technical details, one on managerial details and one recommending task assignment.

To study these tasks we used an internal bug database. This database is used for bug reporting, tracking and resolution as well as general task and feature tracking across the company. We expect ChipNeMo models to perform well on this task as a large amount of bug data was included in the DAPT dataset. Additionally, we built a domain-specific SFT dataset for this task that includes examples of the bug summarizing and task assignment tasks.

Often, bug descriptions contain large snippets of log files or code dumps along with long comment histories. In such cases, the bug text is too large for our LLM context windows. To work around this, we implemented two solutions. First, we found and replaced long path names with shorter aliases to allow the model to associate paths that occur in multiple places in the bug without needing to process the entire string. Second, we split the summarization task into an incremental task where the model is tasked with accumulating data across multiple summary and bug data chunks. We use a hierarchical approach where the bug is first separated into chunks that fit into the context window. Those chunks are then summarized and the summaries are accumulated then separated into chunks. This process is repeated until the entire set of summaries fits into a single context window and a single summary is generated. We use this same approach independent of the LLM used for summarization.

\clearpage
\subsection{Examples}
\label{appendix:examples}
\raggedright
\subsubsection{\textbf{Engineering Assistant Chatbot}}
\label{sec:question_design_chat_example}



\begin{tcolorbox}[width=1.0\linewidth, halign=left, colframe=black, colback=white, boxsep=0.01mm, arc=1.5mm, left=2mm, right=2mm, boxrule=0.5pt]\footnotesize

\textbf{Question:}\\ 
Write a System Verilog assert to check if signal A is true within 5 cycles of signal B being true

\vspace{0.2in}
\textbf{Answer:}\\
Sure, here is the System Verilog assert to check if signal A is true within 5 cycles of signal B being true:
\begin{minted}[breaklines]{verilog}
assert property (@(posedge clk) disable iff (~rst_n)
    (signal_B && !signal_A) |-> ##[1:5] signal_A
);
 \end{minted}
This assert checks that if signal B is true and signal A is false, then signal A will become true within the next 5 cycles. The `\#\#[1:5]` syntax is used to specify a delay range from 1 to 5 cycles. The `disable iff ($\sim$rst\_n)` clause ensures that the assert is not triggered during reset.

\end{tcolorbox}

\begin{tcolorbox}[width=1.0\linewidth, halign=left, colframe=black, colback=white, boxsep=0.01mm, arc=1.5mm, left=2mm, right=2mm, boxrule=0.5pt]\footnotesize
\textbf{Question:}\\
why do we add retiming stages?

\vspace{0.2in}
\textbf{Answer:}\\ 
Retiming stages are added to improve the timing of the design and to account for the physical distance between different parts of the chip. By adding retiming stages, we can reduce the amount of time it takes for signals to travel between different parts of the chip, which can help to improve the overall performance of the design. Additionally, retiming stages can also help to reduce the amount of noise and interference that can occur on the signals, which can further improve the performance of the design.
\end{tcolorbox}

\begin{tcolorbox}[width=1.0\linewidth, halign=left, colframe=black, colback=white, boxsep=0.01mm, arc=1.5mm, left=2mm, right=2mm, boxrule=0.5pt]\footnotesize
\textbf{Question:}\\
How do I run multiple seeds of random tests using testgen framework?

\vspace{0.2in}
\textbf{Answer:}\\ 
You can run multiple seeds of random tests using the Testgen framework by using the $rand\_tgen$ target. This target allows you to specify a variety of options for the stopping criteria of random runs using the same Testgen testlist(s). \\
\begin{minted}[breaklines]{verilog}
nvmk -c . rand_tgen TESTLIST=my_testlist MON=1 TGEN.LOOP=50

\end{minted}
This command will run 50 iterations of the $my\_testlist$ testlist, each with a different seed. You can specify the number of iterations and the seed mode using the TGEN.LOOP and SEED options, respectively.
 
\end{tcolorbox}

\vspace*{0.3in}
\subsubsection{\textbf{EDA Script Generation}} 
Some function names and commands are obfuscated.
\label{sec:question_script_example}


\begin{tcolorbox}[width=1.0\linewidth, halign=left, colframe=black, colback=white, boxsep=0.01mm, arc=1.5mm, left=2mm, right=2mm, boxrule=0.5pt]\footnotesize
\textbf{Question:}\\
write a code in $TOOL1$ to print the number of flop cells in the given rectangle (0,0,100,100)

\vspace{0.2in}
\textbf{Answer:}\\ 
\begin{minted}[breaklines, fontsize=\footnotesize]{python}
# Sure, here is a TOOL1 code to print the number of flop cells in the given rectangle (0,0,100,100):

# This code gets the top_ref of the design
top_ref = env().get_top_reference()
# This code gets the cells of the rectangle in the top_ref
cells = env().layout().cells_in_rectangle( top_ref, rectangle(0, 0, 100, 100), hierarchical)
# This code gets the cells that are flop cells
flop_cells = cells.type("is_flop_cell")
# This code prints the number of flop cells
print(flop_cells.size())
\end{minted}
\end{tcolorbox}
\begin{tcolorbox}[width=1.0\linewidth, halign=left, colframe=black, colback=white, boxsep=0.01mm, arc=1.5mm, left=2mm, right=2mm, boxrule=0.5pt]\footnotesize
\textbf{Question:}\\
Write a $TOOL2$ tcl proc to find intersecting logic from pins '\$pinA'  to pin '\$pinB'.

\vspace{0.2in}
\textbf{Answer:}\\ 
\begin{minted}[breaklines, fontsize=\footnotesize]{tcl}
#Sure, here is the code:
proc myfunc {pinA pinB} {
    # Get all the pins in the fanout of the given pin
    set fanout_pins [fanout -from $pinA]
    # Get all the pins in the fanout of the given pin
    set fanin_pins [fanin -to $pinB]
    # Get the intersection of the two collections
    set intersection [common_logic $fanout_pins $fanin_pins]
    # Return the intersection
    return $intersection
}

\end{minted}

\end{tcolorbox}

\onecolumn
\subsubsection{\textbf{Bug Summary and Analysis}} 
Usernames, chip names and paths are obfuscated.
\label{sec:question_design_bugs_example}

\begin{tcolorbox}[width=1.0\linewidth, halign=left, colframe=black, colback=white, boxsep=0.01mm, arc=1.5mm, left=2mm, right=2mm, boxrule=0.5pt]\footnotesize
\textbf{Prompt:}\\
BugId : 002 

Synopsis : Request power/current measurement on fabrics using fabric test module 

Module : Silicon - DFT

Description :

Hi all,

We want to evaluate silicon current (power) consumed by fabric macros in a chip. This will help us correlate this power to pre-Si power estimates.

Here are the conditions we want to look at for this data:

Fabric test module activity: 0/25/50/75/100\% 

Corners: 

typical\_low\_voltage 

typical\_high\_voltage

We want to isolate fabric current (power) as much as possible, so we want to enable only the logic needed for fabric test modules and the fabrics themselves.

If this information can be broken down per module (there are 6 in the chip), that will help.

Severity : 5-Performance 

Priority : 5-Normal

Comments :  

\#1 : Author : User A 

Is this a requirement only for current or for future chips as well ? 

\#2 : Author : User B 

There will likely be interest in future chips to perform this exercise as we push the boundaries of fabric power density for better area. 

\#3 : Author : User A 

What is needed in this bug ?

@User B It is not clear. 

Can you list the specific needs with timelines and priority \\

\vspace{0.2in}
\textbf{Technical Summary:}\\ 

In the internal bug management system, a bug has been filed regarding the power and current measurements of the fabrics in the latest chip. The request is to evaluate the silicon current (power) consumed by the fabric macros in the latest chip and to correlate this power to pre-Si power estimates. The conditions for the measurements are the fabric test module activity at 0/25/50/75/100\% and the corners are typical\_low\_voltage and typical\_high\_voltage. The data needs to be broken down per module (there are 6 in the latest chip). Only the logic needed for fabric test modules and the fabrics themselves need to be enabled. \\
\vspace{0.2in}
\textbf{Managerial Summary:}\\ 
User A asks User B if the need for evaluating the silicon current of fabric macros in a chip is only for that specific chip or for future ones as well. User B replies that there will likely be interest in future chips to perform this exercise as they push the boundaries of fabric power density for better area. User A then asks what is needed in this bug and asks for specific needs with timelines and priority. \\
\vspace{0.2in}
\textbf{Task Assignment:}\\
User B

\end{tcolorbox}

\end{document}